%% file: acl_latex.tex
\documentclass[11pt]{article}

\usepackage[preprint]{acl}

\usepackage{times}
\usepackage{latexsym}

\usepackage[T1]{fontenc}

\usepackage[utf8]{inputenc}

\usepackage{microtype}

\usepackage{inconsolata}

\usepackage{graphicx}

\usepackage{booktabs}
\usepackage{multirow}
\usepackage{array}
\usepackage{tikz}
\usetikzlibrary{positioning,arrows.meta,shapes,fit,backgrounds,calc}
\usepackage{xcolor}
\usepackage{colortbl}
\usepackage{enumitem}
\usepackage{pifont}
\usepackage{adjustbox}
\usepackage{algorithm}
\usepackage{algpseudocode}
\usepackage[capitalize,noabbrev]{cleveref}
\usepackage{authblk}  %
\usepackage{tcolorbox}

\definecolor{promptbarbg}{gray}{0.27}
\newtcolorbox{promptbox}[1]{%
  colback=white,
  colframe=promptbarbg,
  colbacktitle=promptbarbg,
  coltitle=white,
  fonttitle=\bfseries,
  halign title=center,
  title={#1},
  boxrule=0.5pt,
  arc=3pt,
  left=6pt, right=6pt, top=4pt, bottom=4pt,
}

\newif\ifshowcomments

\newcommand{\DeclareInlineComment}[3]{%
  \expandafter\newcommand\csname #1\endcsname[1]{%
    \ifshowcomments\textcolor{#2}{\textbf{[#3:}\,##1\textbf{]}}\fi%
  }%
}

\DeclareInlineComment{eh}{blue!70!black}{Eran}
\DeclareInlineComment{dw}{cyan!70!black}{David}
\DeclareInlineComment{hw}{magenta!70!black}{Han}
\DeclareInlineComment{esc}{purple}{Elias}
\DeclareInlineComment{mbc}{orange}{Mohit}
\DeclareInlineComment{id}{olive!70!black}{Ido}
\DeclareInlineComment{todo}{green!70!black}{Todo}

\title{Who is the Agent to Blame?\\ Localizing Faithfulness and Citation Mistakes in Agentic Deep Research}

\author{
    \textbf{Eran Hirsch$^{1}$} \qquad
    \textbf{David Wan$^{2}$} \qquad
    \textbf{Han Wang$^{2}$} \\[0.7em]
    \textbf{Elias Stengel-Eskin$^{3}$} \qquad
    \textbf{Mohit Bansal$^{2}$} \qquad
    \textbf{Ido Dagan$^{1}$}
}
\affil{$^{1}$Bar-Ilan University \qquad{} $^{2}$UNC Chapel Hill \qquad $^3$University of Texas at Austin}
\affil{\tt eran.hirsch@biu.ac.il \quad{} \{davidwan, hwang, mbansal\}@cs.unc.edu \\ \tt \quad esteng@utexas.edu \quad{} dagan@cs.biu.ac.il} 

\begin{document}
\maketitle
\begin{abstract}
Deep research (DR) systems produce long-form cited reports by orchestrating multiple agents that search and synthesize information from the web. Citations are the primary mechanism for evaluating the faithfulness of these reports, yet current DR systems exhibit poor citation recall. Moreover, improving citation recall is challenging because DR systems are complex multi-agent architectures where information passes through agents like a telephone game, and both content and citations can get corrupted along the way. We propose an evaluation method that pinpoints which agent introduced each error by locally testing agent invocations for faithfulness and verifiability relative to their own inputs. Furthermore, we propose a four-type taxonomy to categorize the discovered errors: hallucination, uncited input reliance, uncited output, or insufficient citations. Applying our method to three top-ranked open-source DR systems, we obtain actionable diagnostics.  Almost every agent makes a lot of mistakes with the exception being those that summarize a single document. We find that the dominant error type varies systematically across agents, where the orchestrator mistakes are mostly citation-related. We find that 84.7\% of final-report errors in AI-Q originate at the orchestrator, roughly 31\% of them hallucinations and the rest citation mistakes. Guided by these insights, we demonstrate that two simple interventions raise citation recall by 5\% without degrading output quality.\footnote{Code available at \url{https://github.com/eranhirs/who-is-the-agent-to-blame} }

\end{abstract}

\begin{figure*}[t]
\centering
\includegraphics[width=\textwidth]{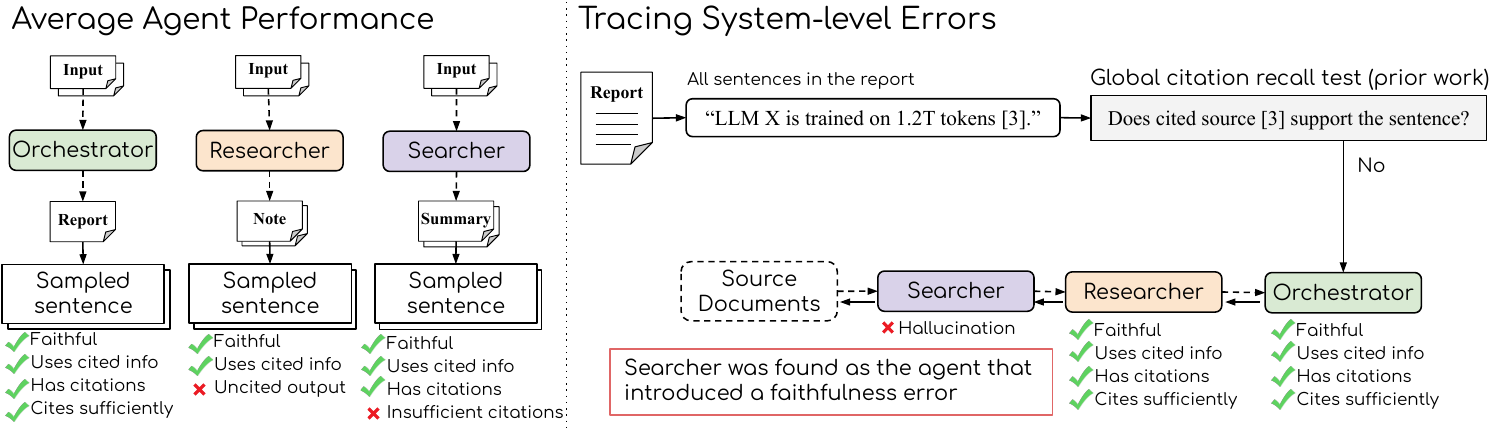}
\caption{Our two evaluation goals for DR systems (\cref{sec:framework}). \textbf{Left:} Our framework samples sentences per agent to produce statistics on how many mistakes it makes, and of what kind. \textbf{Right:} For each sentence in the final report, our framework localizes citation recall errors to the agent that introduced them and classifies errors by type. Dashed arrows represent the original information flow in the DR system; solid lines represent our evaluation process.
}
\label{fig:overview}
\end{figure*}

\section{Introduction}
\label{sec:introduction}

Deep research (DR) systems autonomously search the web and synthesize long-form cited reports in response to a user request.
Similar to how humans perceive the factuality of online text \citep{redi-etal-2019-citation_needed}, citations are a primary mechanism for verifying and measuring the factuality of DR reports \citep{han-etal-2026-deer, patel-etal-2025-deepscholarbench, skarlinski-etal-2024-paperqa2, du-etal-2025-drb}, making citation reliability a central design concern for developers of DR systems.
The standard metric for citation quality, \textit{citation recall} -- the fraction of claims supported by their cited sources \citep{zhang-etal-2025-longcite,gao-etal-2023-alce} -- reveals that current DR systems perform poorly \citep{han-etal-2026-deer,patel-etal-2025-deepscholarbench}.

However, improving citation recall is challenging, particularly with current complex DR system architectures \citep{2024autosurvey,li2025webweaverstructuringwebscaleevidence,tongyidr,asai2026synthesizing}, where information passes through multiple agents, like in a ``telephone game'' (\cref{fig:dr_systems_information_flow}). As content crosses agent boundaries, and often gets compressed, both core information as well as citations may get corrupted \citep{laban2026llmscorruptdocumentsdelegate}.  

Aiming to provide DR developers with a better understanding of citation recall errors and their sources, we propose a novel evaluation methodology that addresses two primary goals.
First, we evaluate the citation performance of individual agent invocations, with respect to their own inputs. This allows two kinds of evaluation: (a) measuring the average citation performance of each agent component in the DR system, as well as (b) tracing which individual agents are responsible for the eventual citation errors that are present in the final DR report.
Second, when an agent makes a mistake, we identify the type of error made; this could be a \emph{hallucination} \citep[not supported by any source; ][]{he2025precise}, relying on input that was itself uncited in a prior agent step (\emph{uncited input reliance}), carrying no citations to the current agent's own output (\emph{uncited output}), or having citations that do not fully cover its sources (\emph{insufficient citations}).

We apply our framework to three top-ranked open-source DR systems, Nvidia AI-Q \citep{nvidia2026aiq}, MS-Agent \citep{modelscope2026msagent}, and TrajectoryKit \citep{trajectorykit2026}. We use examples from the dominant DeepResearch Bench \citep{du-etal-2025-drb} and obtain informative findings. For example, at the agent level, we find that most agents make a lot of mistakes, except those that summarize a single document. Also, that in the two stronger-model systems the dominant failure mode shifts from hallucination at the shallower agents to citation errors at the orchestrator, while the smaller-model system hallucinates at every stage. At the system level, 84.7\% of final-report errors in AI-Q are attributed to the orchestrator (main) agent, where the majority of these are citation-related and a sizable 31\% of the orchestrator's own errors are hallucinations. 
To illustrate the usefulness of such diagnostics, we demonstrate two simple interventions: (1) instruct the orchestrator to avoid using uncited information, and (2) replace synthesized researcher notes with raw source snippets. These interventions successfully raise citation recall by 5\% and citation precision by 3\% to 7\% percentage points, without reducing output quality.

Overall, our contributions include:
\begin{enumerate}
\item A formal, human-validated method for evaluating a single agent invocation, which locally tests faithfulness and verifiability and enables both agent-level and system-level analyses (\cref{fig:overview}).
\item A four-type error taxonomy that distinguishes failure modes across agent boundaries.
\item An empirical study revealing prominent error sources and types in three recent competitive open DR systems
(\cref{sec:agent_analysis,sec:tracing_analysis}).
\item A demonstration of two targeted interventions, guided by our diagnostics, which substantially improve citation quality (\cref{sec:application}).
\end{enumerate}

\section{Background}
\label{sec:background}

\begin{figure}[t]
\includegraphics[width=\columnwidth]{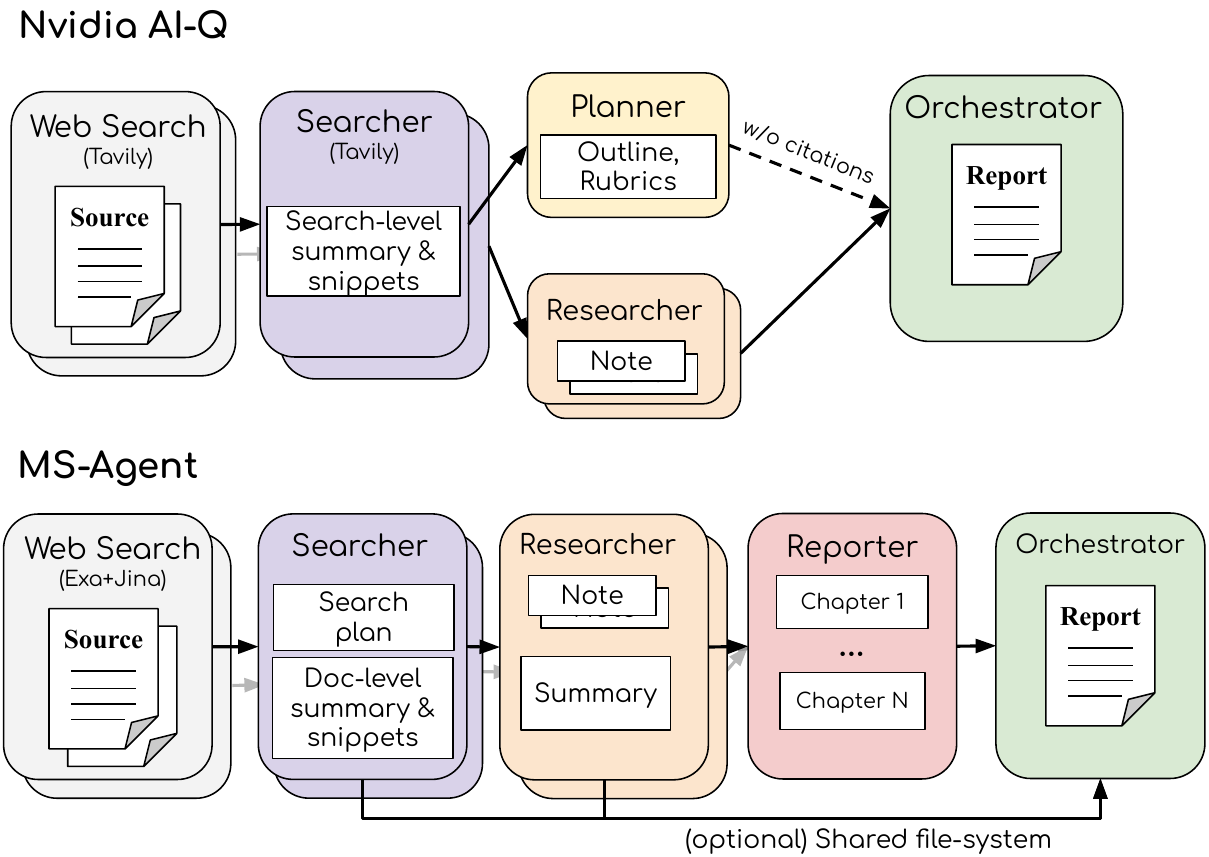}
\centering
\caption{Abstract information flow scheme of two state-of-the-art open-source DR systems, Nvidia AI-Q \citep{nvidia2026aiq} and MS-Agent \citep{modelscope2026msagent}. Each color represents a different agent type and each colored block represents a different agent invocation. Both systems share the same logical agents of searchers, researchers, and an orchestrator, as is common in open-source deep research systems (see \Cref{sec:background}). All systems, including TrajectoryKit \citep{trajectorykit2026}, which we also evaluate, are described in detail in \cref{appendix:system_specific_logic}.}
\label{fig:dr_systems_information_flow}
\end{figure}

DR systems are multi-agent systems that orchestrate web search,
summarization, and long-form report writing.
\Cref{fig:dr_systems_information_flow} contrasts two open-source
instances: Nvidia AI-Q \citep{nvidia2026aiq} and MS-Agent \citep{modelscope2026msagent}. These systems are ranked \#1 and \#2, respectively, in the dominant DeepResearch Bench open-source leaderboard \citep{du-etal-2025-drb} as of May 2026. Despite implementation differences, they share logical roles: a \emph{web search engine} that retrieves source pages; a \emph{searcher} that compresses the retrieved content into doc- or search-level summaries and snippets; a \emph{researcher} that writes natural-language notes with citations, based on the searcher's output; and an \emph{orchestrator} that assembles content into the final cited report.
Beyond these shared roles, AI-Q has a \emph{planner} agent that creates the outline and rubrics, while MS-Agent has the orchestrator write the plan before any web search has been performed. Another difference is that MS-Agent has a \emph{reporter} writing one chapter at a time. Both systems have an iterative phase where the orchestrator edits the final report.
The two systems also differ in the model assigned to each role: AI-Q pairs a frontier model at the planner and orchestrator with a small open-weights model at the researcher, whereas MS-Agent grades model capability with the depth of the agent, from GPT-5 nano at the searcher to GPT-5 mini at the reporter and GPT-5 at the orchestrator (\cref{tab:agent_llms}). This heterogeneity turns out to matter for the errors each agent makes (\cref{sec:agent_analysis}).
We additionally evaluate TrajectoryKit \citep{trajectorykit2026}, ranked \#3 on the same leaderboard, which instantiates the same logical roles. Unlike the other two systems, TrajectoryKit runs its agents on a substantially smaller open-weights model \texttt{gpt-oss-20b} \citep{openai2025gptoss120bgptoss20bmodel}, so its model capability does not vary with the depth of the agent.
Other deep research systems share similar design principles \citep{li2025webweaverstructuringwebscaleevidence, li2025webthinker}.
Across all of these architectures, the telephone-game failure modes described in \cref{sec:introduction} can arise at many points: when information crosses between agents, when an agent's history is truncated or summarized, and even between subsequent writing steps within a single agent. Our work develops evaluation methodologies to trace these errors.

The hallucination problem in LLMs motivated a growing body of work on attributed text generation \citep{gao-etal-2023-alce, han-etal-2026-deer, zhang-etal-2025-longcite}, where output sentences are paired with supporting citations. In DR specifically, citation recall is the standard factuality metric for the entire report \citep{han-etal-2026-deer, patel-etal-2025-deepscholarbench, du-etal-2025-drb,li-etal-2026-drb2}. For each sentence, citation recall checks whether a citation is needed \citep{zhang-etal-2025-longcite} and, if so, whether the associated citations support the claim. Our work goes beyond aggregate citation recall by differentiating specific error types and tracing them to the responsible agent.

\section{Evaluation Framework}\label{sec:framework}

We propose a framework to assist DR developers in identifying the weak parts of their system design in terms of faithfulness and citation mistakes, as measured by citation recall. Our approach is to provide error categorization and localization, measuring what kind of errors occurred and which agent is to blame. 

The building block of our evaluation framework is a method to evaluate the output of a single agent invocation with respect to the inputs that it received (\cref{sec:method}). In contrast to the standard method for evaluating citation recall \citep{gao-etal-2023-alce} which \emph{globally} evaluates sentences in the final output report and their citations against the source documents, our method \emph{locally} evaluates the output of an agent invocation relative to its own inputs. This local approach enables us to pinpoint an error to a specific agent. In addition, this local approach makes faithfulness tests more tractable, since we evaluate an output only relative to its inputs, rather than against potentially hundreds of full source documents.

Using this local agent invocation evaluation method, our framework addresses two complementary evaluation goals (depicted in \cref{fig:overview}): the agent-level evaluation, which evaluates an agent's average performance across all of its sampled invocations (\cref{sec:agent_analysis}), and the system-level evaluation, which traces only the errors that reach the final report back to their origin (\cref{sec:tracing_analysis}). These two evaluation types are complementary, because not every output of an agent is eventually used by the final report. In the agent-level evaluation, we report the average percentage of mistakes that the agent makes, as well as an error distribution of those mistakes. In the system-level evaluation, we report the percentage of actual mistakes in the DR report, their origin, and error categories. In analogy to system vs. unit testing, system-level evaluation tells us about the errors that made it to the final report, but can hide mistakes that specific agents made. Agent-level evaluation tells us about the errors that an agent makes, but not whether these errors affected the final report.

\begin{figure*}[t]
\centering
\includegraphics[width=\textwidth]{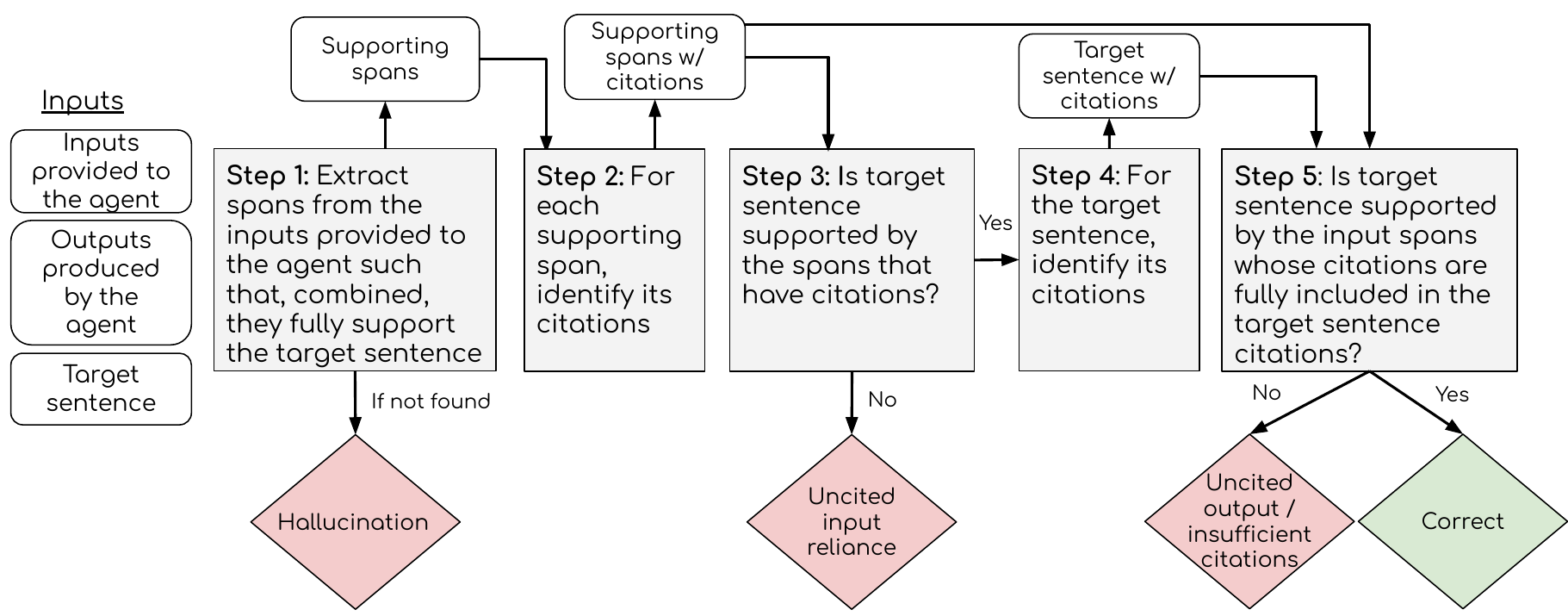}
\caption{Overview of our algorithm for evaluating information errors in a synthesized target sentence within a single agent invocation. Each box corresponds to one step of the method described in \cref{sec:synthesized}. %
}
\label{fig:algorithm-overview}
\end{figure*}

\section{Evaluating an Agent Invocation}
\label{sec:method}

To better understand citation recall mistakes produced by deep research systems (\cref{sec:agent_analysis,sec:tracing_analysis}), we start by describing how to evaluate a single agent invocation. We propose a method designed to identify the correctness of the agent's output text, relative to its inputs, in terms of faithfulness (i.e., no hallucinations relative to the input) and verifiability (i.e., citation markers in the output match the citation markers in the input). 

\subsection{Settings}\label{sec:setting}

The scope of our method is the evaluation of a single agent invocation. We extract all the inputs that the agent received, possibly from many separate LLM invocations (e.g., a single agent invocation can include many web searches). In addition, we extract all the outputs that the agent invocation created, possibly from many LLM invocations (e.g., a single agent invocation can include writing to many files). Examples of such agent invocations are depicted in \cref{fig:dr_systems_information_flow}. This extraction phase requires a different implementation per DR system (\cref{appendix:system_specific_logic}).

From the agent's perspective, it is assumed that the inputs are faithful to the sources, and that each input info unit is supported by the citations associated with it. The method is not concerned with mistakes that were possibly inherited from prior agent invocations.
Accordingly, the agent's responsibility is to produce output info units that are faithful to the cited input info units, while carrying forward all associated citations. If there are some input info units that are not associated with any citations, they should not be utilized by the agent.

\subsection{Evaluation Method}
\label{sec:algorithm}

The inputs to our algorithm are the extracted inputs and outputs of the agent invocation (\cref{sec:setting}). The output of our algorithm is a list of errors that the agent introduced.
We differentiate between two types of agent invocation outputs: \emph{extracted snippets} and \emph{synthesized information}. Extracted snippets are spans of text that appear verbatim in some raw document. Synthesized information units are generated by the agent model to paraphrase, summarize, or consolidate raw documents and snippets. The error categories differ by output type. 

For synthesized outputs, each input and output carries information and, optionally, attributing citation markers (e.g., \texttt{[N]}). Conceptually, the following things can go wrong: (1) the output is unfaithful to the input (\emph{hallucination}); (2) the output is faithful to the input but draws on input that was itself uncited (\emph{uncited input reliance}); (3) the output is faithful to cited input but carries no citations at all (\emph{uncited output}); or (4) the output has citations but fails to include the complete set of citations of the input it relied on (\emph{insufficient citations}).
For snippet outputs, which by definition belong to a single identified document, citation is inherited from the source identity, so the only concern is whether extracting the snippet removed important context, rendering it uninterpretable or hallucinated (\emph{snippet missing context}).
We describe the method for each output type separately: synthesized information in \cref{sec:synthesized} and snippets in \cref{sec:snippets}.

\subsubsection{Evaluation for Synthesized Information}
\label{sec:synthesized}

The algorithm, explained below, is visualized in \cref{fig:algorithm-overview} (see \cref{appendix:algo_pseudocode} for the pseudo-code). Because DR systems typically attach citations at the sentence level, we segment each synthesized output information unit into sentences and run the algorithm on each sentence independently, returning one of the four error labels above or \emph{no mistake}. The algorithm, a sequence of attribution and entailment tasks, proceeds as follows, where each step corresponds to a box in \cref{fig:algorithm-overview}:

\begin{itemize}
    \item \textbf{Step 1 -- Faithfulness test.} An LLM extracts all supporting spans from the information provided by prior agent invocations; spans should be exhaustive, even if redundant or only partially supporting. If the extracted spans do not fully support the target sentence, either because no spans were found or because they support it only in part, this indicates a \emph{hallucination} (terminal verdict). Otherwise, the output is faithful to its inputs and we proceed to the verifiability tests.\footnote{We include two validations for the extracted spans: (a) spans must appear verbatim in the input texts, and (b) an entailment check verifies that the union of spans supports the target sentence. On failure, we retry with self-critique feedback. If the retry budget is exhausted, we record the case as \emph{evaluation failed} and filter it out. }
    \item \textbf{Steps 2--3 -- First verifiability test (uncited input reliance).} Given that the output is faithful, we now check whether it relies on uncited input. Because citations in DR outputs are often not attached to every sentence, an LLM first identifies which citations are assigned to each input span (Step 2). We then remove uncited spans and test whether the remaining cited spans still entail the target sentence (Step 3). If not, the output is classified as \emph{uncited input reliance} (terminal verdict).
    \item \textbf{Steps 4--5 -- Second verifiability test (citation alignment).} Finally, we check whether the target sentence's own citations are complete, in the sense that they carry all necessary citations from the cited input spans. An LLM first identifies the target sentence's citations (Step 4). We then check alignment: if the target sentence carries no citations at all, we conclude it is an \emph{uncited output} (terminal verdict). If it does have citations, we keep only input spans whose citations are a subset of the target's citations and re-run the entailment test. If this subset no longer supports the target sentence, we conclude that the target has \emph{insufficient citations} (terminal verdict). Otherwise the output is correct.%
\end{itemize}

\paragraph{Algorithm implementation.}
All our steps are based on existing prompts from prior work. The Step 1 span extraction prompt is based on the prompt used for Localized Attribution Queries \citep{hirsch-etal-2025-laquer}. The Steps 2 and 4 citation classification is based on the approach introduced by DEER \citep{han-etal-2026-deer} to semantically identify which citations are related to a sentence or span. All other prompts are entailment tests based on the LongCite prompt \citep{zhang-etal-2025-longcite}. The LLM used as the judge is \texttt{gpt-5-mini-2025-08-07} \citep{singh2026openaigpt5card}. 
The full prompts are provided in \cref{appendix:prompts}.

\subsubsection{Evaluation for Extracted Snippets}
\label{sec:snippets}

\citet{zhang-etal-2023-extractive} identified that a snippet extracted from a larger source can remove necessary context, creating uninterpretable or even hallucinated information. Accordingly, we run a single entailment test that asks whether the source document supports the entire snippet. If the test passes, the snippet is considered correct. If not, it receives the error label \emph{snippet missing context}.

\subsection{Human Assessment of Method Implementation Performance}
\label{sec:human_analysis}

In order to assess the reliability of our automated agent invocation evaluation procedure, we asked humans to perform the same classification task as our algorithm in \cref{fig:algorithm-overview}.
Each annotation task showed the annotator a target sentence (in the report context) and the agent invocation's inputs, and asked them to pick one of six labels (\emph{no citation needed}, \emph{hallucination}, \emph{uses uncited info}, \emph{uncited}, \emph{insufficient citations}, \emph{no issues}), the last five corresponding to the method's terminal verdicts for synthesized information. The full annotation guidelines are included in \cref{appendix:annotation_guidelines}.

\paragraph{Setup.}
We sampled 50 sentences from AI-Q examples and had each sentence annotated independently by two annotators drawn from a pool of four. In case of disagreements, a third annotator arbitrated and made the final call.
The annotators were all graduate students with prior experience reading academic literature and using deep research systems. 
Annotators received an orientation session and completed calibration tasks before beginning the main annotation.

\paragraph{Results.}
The inter-annotator agreement before arbitration was Cohen's $\kappa{=}0.71$ (substantial) \citep{Cohen1960ACO, Landis1977TheMO}. We then compared the arbitrated labels with the method's verdicts, yielding $76\%$ exact matches and $\kappa{=}0.62$ (substantial).

In \cref{appendix:span_validation} we validate the supporting spans, using the same annotation effort. The localization process agrees with the annotators on $75\%$ of the relevant sentences. This validates the system-level analysis (\cref{sec:tracing_analysis}), whose recursion follows these spans back from the final report to the agent that introduced the error.

\begin{figure*}[t]
\includegraphics[width=\textwidth]{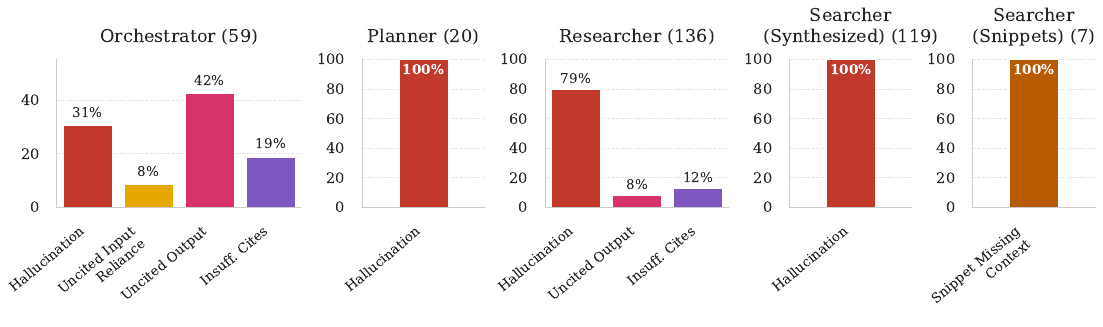}
\centering
\caption{Agent-level error distribution for each agent of the Nvidia AI-Q DR system, showing the breakdown of the error portion from \Cref{tab:agent_level_performance}. The corresponding distributions for the MS-Agent and TrajectoryKit DR systems are in \cref{fig:ms_agent__agent_level_analysis,fig:trajectorykit__agent_level_analysis} (appendix).
}
\label{fig:agent_level_analysis}
\end{figure*}

\section{Evaluating Average Agent Performance}
\label{sec:agent_analysis}

The goal of this evaluation is to extract average performance statistics per agent.
Ideally, we would evaluate all the sentences of all agent invocations per agent, but that would be intractable. Instead, we sample sentences without repetition from all the invocations of each agent type in each example.

For each sentence sampled, we first classify if a citation is needed \citep{redi-etal-2019-citation_needed, zhang-etal-2025-longcite}. If not, we discard it. We use the LongCite ``is citation needed'' prompt (\cref{appendix:prompt-need-citation}).\footnote{Agents that extract snippets are sampled at the invocation level and not the sentence level. They also do not include the is-citation-needed filter.}
For each sample, we then run the agent invocation evaluation method described in \cref{sec:algorithm}. Finally, for each agent, we report its mistake rate and the distribution of error types.

\subsection{Experimental Setup}
We apply our method to the three state-of-the-art open-source deep research systems described in \cref{sec:background}: Nvidia AI-Q \citep{nvidia2026aiq}, MS-Agent \citep{modelscope2026msagent}, and TrajectoryKit \citep{trajectorykit2026}. The LLMs used by each agent are listed in \cref{tab:agent_llms} (appendix). We evaluate on 20 examples from DeepResearch Bench \citep{du-etal-2025-drb}. For each example, we sample up to 10 sentences per agent, yielding roughly 200 sentences per agent, with exact counts presented in \cref{tab:agent_level_performance}.

\subsection{Results}

Per-agent correct and mistake percentages for all three systems are reported jointly in \cref{tab:agent_level_performance}. The per-agent error distribution for AI-Q is shown in \cref{fig:agent_level_analysis}; the corresponding MS-Agent and TrajectoryKit distributions are in \cref{fig:ms_agent__agent_level_analysis,fig:trajectorykit__agent_level_analysis} (placed in the appendix).

\begin{table}[t]
\centering
\begin{adjustbox}{max width=\columnwidth}
\input{auto_generated_figures/agent_level_performance.tex}
\end{adjustbox}
\caption{Agent-level performance for Nvidia AI-Q, MS-Agent, and TrajectoryKit.  $^{*}$For the planner agent, ``uncited output'' is not considered as an error, because this is by design (\cref{fig:dr_systems_information_flow}). 
}
\label{tab:agent_level_performance}
\end{table}

\paragraph{Agents that summarize a single document make the fewest mistakes.}
Almost every agent in the three systems makes significant mistakes on a substantial share of its sampled sentences. The clear exceptions are the searcher components that compress one document at a time: the AI-Q searcher snippets (3.8\%), the MS-Agent searcher snippets and synthesized summaries (6.4\% and 0.9\%, respectively), and the TrajectoryKit searcher synthesized summaries (14.9\%). AI-Q's searcher synthesized summaries also perform a summarization, but over all documents returned for a query rather than one, and their mistake rate rises to 62.0\%. The remaining spread within the single-document group can be attributed to the underlying model rather than the architecture, as TrajectoryKit's \texttt{gpt-oss-20b} digests (14.9\%) are an order of magnitude worse than MS-Agent's otherwise identical per-document summaries (0.9\%).

\paragraph{Deeper agents hallucinate less.}%
Agents deeper in the information flow, which receive information that was synthesized more times, hallucinate less than shallower ones. Only 31\% of the AI-Q orchestrator's mistakes are hallucinations, against 79\% for its researcher. MS-Agent shows the same contrast even more sharply: none of the orchestrator's mistakes are hallucinations, against 78\% for the reporter and 92\% for the researcher. 

A plausible explanation is the capability of the underlying model rather than depth in itself, as the shallower agents run substantially weaker LLMs than the orchestrator in both systems (see \cref{tab:agent_llms} in the appendix). TrajectoryKit is consistent with this explanation: it runs the same model at every stage and correspondingly shows no shift, with hallucinations accounting for most of the mistakes at each of its agents.

\section{Tracing System-level Errors}
\label{sec:tracing_analysis}

We propose a system-level evaluation to trace and localize global citation recall errors to the specific agents that created them and to categorize them into specific error types. We first describe an algorithm for the evaluation (\cref{sec:tracing_algo}) and then discuss the results (\cref{sec:tracing_algo_results}).

\subsection{Tracing Algorithm}\label{sec:tracing_algo}

We first have to find which sentences have citation recall issues globally. Then, for each such sentence, we run multiple local tests recursively, starting from the agent that created the final report, until we identify which agent introduced the error.

\subsubsection{Calculating Citation Recall in DR}

The implementation of the global citation recall in Step 1 follows standard practice \citep{zhang-etal-2025-longcite} of answering two questions: (1) whether the target sentence needs a citation, and (2) whether it has citations that support it. Similar to the local algorithm, we use a citation-identification prompt (drawn from DEER \citep{han-etal-2026-deer}) to semantically find which citations are aligned with the target sentence.
The LLM judge for all prompts is \texttt{gpt-5-mini-2025-08-07} \citep{singh2026openaigpt5card}.

Because DR systems can hallucinate URLs \citep{onweller2026citedverifiedparsingevaluating}, we use the tracing log to build a ground-truth list of retrieved URLs and filter out any hallucinated citations. We also adopt a no-crawl design, extracting raw documents from the tracing log instead of re-crawling, which sidesteps crawl-fragility issues (disagreement between crawlers, paywalls, dynamic content).

\subsubsection{Finding the Agent to Blame}

Given a target sentence in the final report that has global citation recall errors, we now recursively iterate over agent invocations and try to find the agent invocation to blame by running local tests.

We first run the local agent invocation test from \cref{sec:algorithm}, which checks whether the agent's output is faithful to and correctly cited against its inputs. If the test detects an error, we attribute it to this agent and return. Otherwise, we choose one or more targets to recurse into and return their found errors.
Specifically, if the target sentence has no errors, this means that the local algorithm from \cref{sec:algorithm} must have identified spans that fulfill the following conditions: (1) they support the target sentence, (2) they have citations, and (3) their citations are included in the target sentence. 
For each such span, we check if it originates from a raw document, which would indicate that we reached the end of the recursion without finding any agent to blame. Otherwise, we recurse into the agent invocation that produced the span, applying the same procedure to the sentence containing it. The recursion terminates when an error is found or a raw document is reached. The final output is a list of found errors.

\subsection{Results}\label{sec:tracing_algo_results}

\begin{figure}[t]
\includegraphics[width=1\columnwidth]{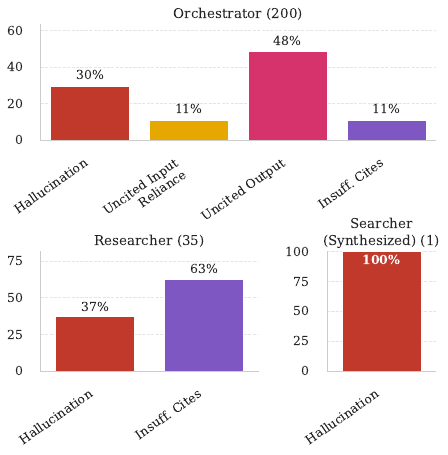}
\centering
\caption{System-level error distribution for Nvidia AI-Q. The corresponding distributions for MS-Agent and TrajectoryKit are in \cref{fig:ms_agent__system_level_verdicts,fig:trajectorykit__system_level_verdicts} (appendix).
}
\label{fig:system_level_verdicts}
\end{figure}

\begin{table}[t]
\centering
\small
\begin{adjustbox}{max width=\columnwidth}
\input{auto_generated_figures/system_level_error_origin_distribution.tex}
\end{adjustbox}
\caption{System-level error-origin distribution for Nvidia AI-Q, MS-Agent, and TrajectoryKit. For each system, the share of confirmed final-report errors attributed to each agent.
}
\label{tab:system_level_error_origin_distribution}
\end{table}

The experimental setup for this evaluation uses the same sample from \cref{sec:agent_analysis}. The global citation recall for AI-Q is 58.7\%, for MS-Agent is 28.5\%, and for TrajectoryKit is 7.1\%. Statistics on agents that were blamed for creating errors are reported jointly in \cref{tab:system_level_error_origin_distribution}. The per-category error distribution for AI-Q is shown in \cref{fig:system_level_verdicts}; the corresponding MS-Agent and TrajectoryKit distributions are in \cref{fig:ms_agent__system_level_verdicts,fig:trajectorykit__system_level_verdicts}.
\Cref{tab:examples} (appendix) provides representative examples for each mistake category produced by our algorithm.

\paragraph{The majority of mistakes start at the orchestrator.}

For AI-Q, 84.7\% of final-report errors were attributed to the orchestrator, for MS-Agent 52.6\%, and for TrajectoryKit 100\%. For the first two systems this is despite the orchestrator having one of the lowest agent-level error rates (30.9\% for AI-Q and 16.8\% for MS-Agent in \cref{tab:agent_level_performance}), which illustrates the value of system-level evaluation on top of agent-level evaluation. TrajectoryKit is the opposite case: its orchestrator is also its weakest agent (92.5\%), so the two levels of evaluation agree.

\paragraph{Final output mistakes are mostly citation-related errors.} For AI-Q 70\% of the mistakes are citation-related mistakes at the orchestrator level, and for MS-Agent 99\%. This is consistent with the results of the agent-level evaluation. For MS-Agent, we find that the final report sometimes contains citations to IDs of the reporter's notes instead of source documents (see \cref{appendix:system_specific_logic}). TrajectoryKit is the exception: only 4\% of its orchestrator errors are citation-related, as discussed below.

\paragraph{Hallucinations are a notable share of AI-Q's final-report errors.} Beyond citation-related errors, a substantial fraction of errors that reach AI-Q's final report are hallucinations. Hallucinations account for roughly 31\% of AI-Q's final-report errors (30\% of the orchestrator's 200 errors and 37\% of the researcher's 35), with most originating at the orchestrator. For MS-Agent, hallucinations are far less of an issue, at roughly 11\%. This is notable because hallucinations are arguably more severe than citation-related errors.

\paragraph{The smaller-model system fails in a different way.} For TrajectoryKit,  95\% of its final-report errors are hallucinations. Manual inspection shows a recurring failure mode in which the orchestrator attaches a fabricated precise statistic to a source that does not state it. This is consistent with its markedly lower citation recall, even though its output quality on the RACE metric \citep{du-etal-2025-drb} is competitive with MS-Agent's. This reinforces the model-capability explanation from the agent-level evaluation (\cref{sec:agent_analysis}): the dominant error type is governed less by the multi-agent architecture, which the three systems largely share, than by the capability of the underlying model.

\paragraph{How errors propagate across agents.} Although the orchestrator is the most common origin, a substantial share of final-report errors starts elsewhere (\cref{tab:system_level_error_origin_distribution}), which means several subagents propagate mistakes received as input up to the final report. \cref{tab:error_chains} in the Appendix walks through three representative chains that illustrate example patterns: a searcher fabricates a claim that the researcher and the orchestrator faithfully propagate; the orchestrator abstracts a researcher's fabricated claim while keeping its citations; and unsupported statistics travel verbatim with a citation that never supported them.

\section{Interventions to Improve Citation Recall}
\label{sec:application}

In this section, we show how insights from our evaluation can help improve DR systems. We demonstrate how simple changes, not implemented by the original developers, can improve citation quality on AI-Q. Future work could include more complex fixes, such as using our evaluation's diagnostics as a corrective feedback signal during generation.

We propose two fixes for AI-Q based on our evaluation's findings. The first fix follows from our average agent performance results: the searcher snippets, each extracted from a single document, are the most reliable artifact in the system (3.8\% mistakes), whereas the researcher notes that currently reach the orchestrator are produced by its least reliable agent (70.8\%) and consolidate several documents at once. 
This motivates the following fix: in the orchestrator inputs, we replace the synthesized researcher agents' notes with the snippets of the URLs they cite, so that each unit the orchestrator reads is a single-document extract rather than a multi-document synthesis. This enables the orchestrator to directly access the raw document snippets, and we expect that it will reduce the number of citation-related errors.
For the second fix, we focus on our finding that the final output mistakes are mostly citation-related errors. Accordingly, we append to the orchestrator prompt the following instruction: ``do NOT use information which is not backed by citations.'' We expect that this will have an effect both on the reliance on uncited input, as well as on the uncited output. We evaluate each of these two changes independently.

\paragraph{Experimental setup.} We evaluate on 50 examples from DeepResearch Bench, covering all of its English queries, and report the RACE quality metric \citep{du-etal-2025-drb}. In addition, we report the standard citation recall metric, as described in \cref{sec:tracing_analysis}. Finally, we also calculate citation precision using the citation precision prompt from LongCite \citep{zhang-etal-2025-longcite}. 

\paragraph{Results.} \cref{tab:aiq-intervention} shows that for AI-Q, both interventions substantially increase citation recall by 5\% without significantly changing output quality. Both also increase citation precision by 3\% to 7\%. The two interventions perform comparably on recall.

To further analyze the citation guidance intervention, we run our system-level tracing method (\cref{tab:avoid_uncited_error_origin}; \cref{fig:avoid_uncited_system_level_verdicts} in the appendix). We find that \textbf{the orchestrator's share of errors drops from 84.7\% to 77.0\% after the intervention}.
In addition, we find that the orchestrator's \textbf{citation-related errors after the intervention drop from 70\% to 60\%}, compared to the citation-related errors made before the intervention (\cref{fig:system_level_verdicts}). Overall, our intervention was able to reduce orchestrator citation-related error as expected, resulting in a stronger DR system across all metrics. Future work could further attempt to reduce orchestrator errors, or divert the focus to fixing researcher-induced errors.

\begin{table}[t]
\centering
\small
\begin{adjustbox}{max width=\columnwidth}
\input{auto_generated_figures/aiq_intervention}
\end{adjustbox}
\caption{Results for simple AI-Q DR system modifications based on our evaluation findings. Values after $\pm$ are standard errors of the mean.}
\label{tab:aiq-intervention}
\end{table}

\begin{table}[t]
\centering
\begin{adjustbox}{max width=\columnwidth}
\input{auto_generated_figures/system_level_error_origin_distribution_avoid_uncited.tex}
\end{adjustbox}
\caption{System-level error-origin distribution for AI-Q after the citation guidance prompt change. Compare against \cref{tab:system_level_error_origin_distribution}: the orchestrator's share decreases from 84.7\% to 77.0\%, while the researcher's share increases from 14.8\% to 23.0\%.}
\label{tab:avoid_uncited_error_origin}
\end{table}

\section{Conclusion}

We presented a methodology for diagnosing citation recall errors in deep research systems along two complementary axes: what kind of error occurred, and which agent introduced it. Our central finding is that the dominant error type varies systematically across agents: in the two stronger-model systems, the orchestrator hallucinates less and its mistakes are mostly citation-related, whereas the smaller-model system fabricates content at every stage. At the system level, the orchestrator, despite having one of the lowest agent-level error rates, is responsible for 84.7\% of final-report errors in AI-Q. These diagnostics translate directly into targeted fixes: even a one-sentence prompt intervention raises citation recall by 5\% and citation precision by 3--7\% without reducing output quality. A natural follow-up is post-hoc mitigation that feeds the system-level verdicts back into the DR loop as a critique signal, and we expect this approach to generalize to other multi-agent attributed generation settings.

\section*{Limitations}

\paragraph{LLM-as-a-judge reliability.}
The faithfulness and verifiability checks of our algorithm (\cref{sec:method}) rely on LLM-as-judge prompts adapted from prior work. We mitigate this concern in two ways: first, all checks are attribution and entailment tests drawn from prior work (\cref{sec:algorithm}); second, a human annotation study (\cref{sec:human_analysis}) shows substantial agreement between annotator consensus and algorithm verdicts.

\paragraph{Attributing error types to model capability.}
Our observation that the dominant error type tracks the capability of the model an agent runs (\cref{sec:agent_analysis,sec:tracing_algo_results}) is observational rather than controlled. Isolating the factor would require running the same system with the same model across roles, which we leave to future work.

\paragraph{Scope of evaluated content.}
We evaluate citations on running-text sentences only. Tabular content (numeric cells with trailing citation markers) is excluded from both the recall headline and the system-level analysis.

\section*{Ethical Considerations}

Our work contributes to the goal of improving faithfulness and verifiability of deep research systems, advocating for better evaluation methods. However, errors in the different steps of the proposed methodology can mislead developers into drawing incorrect conclusions about the quality of the deep research system, either overstating or understating its reliability.

In the human evaluation (\cref{sec:human_analysis}), each annotation task took about 10 minutes on average, for roughly 6 hours of work per annotator. Annotators were paid \$35 per hour, which we consider adequate compensation for this expertise level and country of residence.

AI-writing tools assisted the writing of the paper to improve clarity and coherence. Any writing change was carefully requested from the AI writing tool and edited by the authors before incorporating it into the paper.

\section*{Acknowledgments}

This work was supported in part by the Israel Science Foundation (grants no. 2827/21 and 3182/25), NSF-AI Engage Institute DRL2112635, NSF-CAREER Award 1846185, and a Google PhD Fellowship.
The views contained in this article are those of the authors and not of the funding agency.
Special thanks to Jan Buchmann, Shmuel Amar, Uri Katz, and Aviya Maimon for their valuable discussions at various stages of the project.

\bibliography{custom}

\appendix
\Crefname{appendix}{Appendix}{Appendices}

\section{Algorithm Pseudo-code}
\label{appendix:algo_pseudocode}

\Cref{alg:writing-step-eval} provides the pseudo-code for the single agent invocation evaluation procedure described in \cref{sec:method}. \Cref{alg:snippet-eval} provides the simpler procedure used for snippet outputs. The helper routines \textsc{ExtractSupportingSpans} (Step 1), \textsc{CitationsOf} (Steps 2 and 4), and \textsc{Entails} (Steps 1, 3, and 5) are LLM calls; see \cref{appendix:prompts} for the full prompts.

\begin{algorithm*}[t]
\caption{Evaluating a single agent invocation on a target sentence.}
\label{alg:writing-step-eval}
\begin{algorithmic}[1]
\Require Target sentence $s$; input info units $I$ available to the agent invocation (already extracted from the trace; see \cref{appendix:system_specific_logic}); retry budget $K$.
\Statex \textit{Step 1: Faithfulness test (with self-critique retries).}
\State $\textit{fb} \gets \emptyset$ \Comment{feedback for retries}
\For{$k = 1, \ldots, K$}
    \State $S \gets \textsc{ExtractSupportingSpans}(s, I, \textit{fb})$
    \If{$S = \emptyset$} \Return \textsc{hallucination} \EndIf
    \If{\textbf{not} $\textsc{AllVerbatim}(S, I)$}
        \State $\textit{fb} \gets$ ``some spans are not verbatim''; \textbf{continue}
    \EndIf
    \If{$\textsc{Entails}\!\left(\bigcup S,\, s\right)$} \textbf{break} \Comment{output is faithful}
    \EndIf
    \State $\textit{fb} \gets$ ``union of spans does not entail $s$''
\EndFor
\If{retries exhausted without success} \Return \textsc{eval\_failed} \EndIf
\Statex \textit{Step 2: Input spans citation classification.}
\State $C_\sigma \gets \textsc{CitationsOf}(\sigma)$ \textbf{for each} $\sigma \in S$
\Statex \textit{Step 3: Uncited input reliance test.}
\State $S^{\mathrm{cited}} \gets \{\, \sigma \in S \,:\, C_\sigma \neq \emptyset \,\}$
\If{\textbf{not} $\textsc{Entails}\!\left(\bigcup S^{\mathrm{cited}},\, s\right)$} \Return \textsc{used\_uncited} \EndIf
\Statex \textit{Step 4: Target sentence citation classification.}
\State $C_s \gets \textsc{CitationsOf}(s)$
\Statex \textit{Step 5: Citation alignment.}
\If{$C_s = \emptyset$} \Return \textsc{uncited} \EndIf
\State $S^{\mathrm{aligned}} \gets \{\, \sigma \in S^{\mathrm{cited}} \,:\, C_\sigma \subseteq C_s \,\}$
\If{\textbf{not} $\textsc{Entails}\!\left(\bigcup S^{\mathrm{aligned}},\, s\right)$} \Return \textsc{insufficient\_citations} \EndIf
\State \Return \textsc{correct}
\end{algorithmic}
\end{algorithm*}

\begin{algorithm}[t]
\caption{Evaluating a snippet output unit.}
\label{alg:snippet-eval}
\begin{algorithmic}[1]
\Require Snippet $o$ with source URL $u$; raw document $d_u$ retrieved for $u$ in the log $L$.
\If{\textbf{not} $\textsc{Entails}(d_u,\, o)$} \Return \textsc{snippet\_missing\_context} \EndIf
\State \Return \textsc{correct}
\end{algorithmic}
\end{algorithm}

\section{Prompts}
\label{appendix:prompts}

This appendix lists the prompts used in the algorithm of \cref{sec:algorithm}, the agent-level analysis of \cref{sec:agent_analysis} and the system-level analysis of \cref{sec:tracing_analysis}.
\Cref{appendix:dr-mods} then briefly describes the modifications applied to AI-Q for the application study (\cref{sec:application}); we omit the full third-party prompts and only describe the changes.

\subsection{Step 1: Span Extraction}
\label{appendix:prompt-extract}

\begin{figure*}[t]
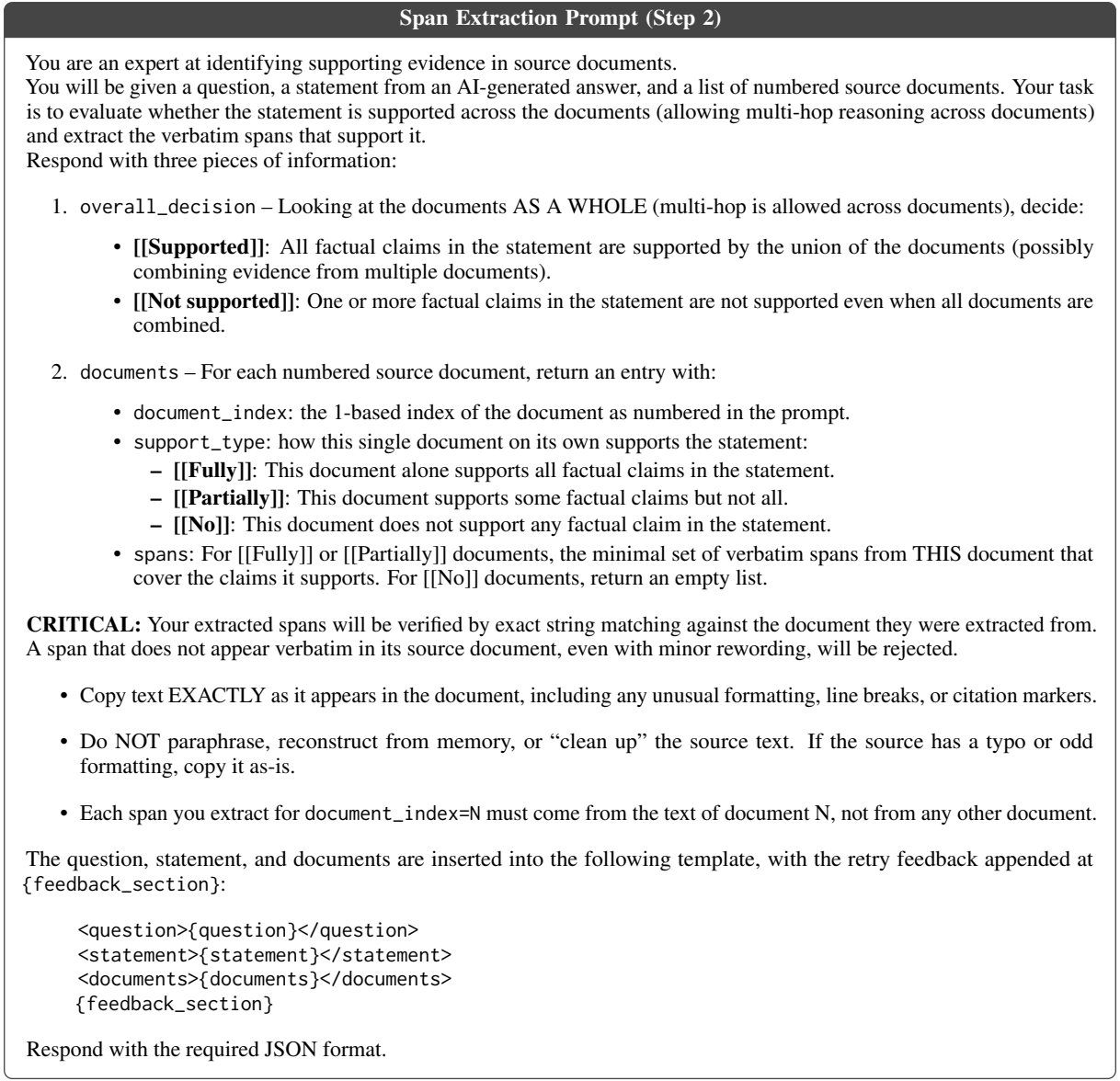

\footnotesize
\begin{promptbox}{Span Extraction Prompt (Step 2)}
You are an expert at identifying supporting evidence in source documents.

You will be given a question, a statement from an AI-generated answer, and a list of numbered source documents. Your task is to evaluate whether the statement is supported across the documents (allowing multi-hop reasoning across documents) and extract the verbatim spans that support it.

Respond with three pieces of information:
\begin{enumerate}
  \item \texttt{overall\_decision} -- Looking at the documents AS A WHOLE (multi-hop is allowed across documents), decide:
  \begin{itemize}
    \item \textbf{[[Supported]]}: All factual claims in the statement are supported by the union of the documents (possibly combining evidence from multiple documents).
    \item \textbf{[[Not supported]]}: One or more factual claims in the statement are not supported even when all documents are combined.
  \end{itemize}
  \item \texttt{documents} -- For each numbered source document, return an entry with:
  \begin{itemize}
    \item \texttt{document\_index}: the 1-based index of the document as numbered in the prompt.
    \item \texttt{support\_type}: how this single document on its own supports the statement:
    \begin{itemize}
      \item \textbf{[[Fully]]}: This document alone supports all factual claims in the statement.
      \item \textbf{[[Partially]]}: This document supports some factual claims but not all.
      \item \textbf{[[No]]}: This document does not support any factual claim in the statement.
    \end{itemize}
    \item \texttt{spans}: For [[Fully]] or [[Partially]] documents, the minimal set of verbatim spans from THIS document that cover the claims it supports. For [[No]] documents, return an empty list.
  \end{itemize}
\end{enumerate}

\textbf{CRITICAL:} Your extracted spans will be verified by exact string matching against the document they were extracted from. A span that does not appear verbatim in its source document, even with minor rewording, will be rejected.
\begin{itemize}
  \item Copy text EXACTLY as it appears in the document, including any unusual formatting, line breaks, or citation markers.
  \item Do NOT paraphrase, reconstruct from memory, or ``clean up'' the source text. If the source has a typo or odd formatting, copy it as-is.
  \item Each span you extract for \texttt{document\_index=N} must come from the text of document N, not from any other document.
\end{itemize}

The question, statement, and documents are inserted into the following template, with the retry feedback appended at \texttt{\{feedback\_section\}}:
\begin{quote}\ttfamily
<question>\{question\}</question>\\
<statement>\{statement\}</statement>\\
<documents>\{documents\}</documents>\\
\{feedback\_section\}
\end{quote}
Respond with the required JSON format.
\end{promptbox}
\caption{Step 2 span-extraction prompt, adapted from the Localized Attribution Queries prompt of \citet{hirsch-etal-2025-laquer}. The \texttt{\{feedback\_section\}} slot is populated by the templates in \cref{fig:prompt-extract-feedback} on retry attempts.}
\label{fig:prompt-extract}
\end{figure*}

\Cref{fig:prompt-extract} shows the prompt used for the LLM call in Step 1 of \cref{sec:method}. Given a target sentence and the documents output by the parent agent invocations, it extracts verbatim supporting spans and emits both a per-document support verdict and an overall verdict. The prompt is adapted from the Localized Attribution Queries prompt of \citet{hirsch-etal-2025-laquer}.

When verification of the extracted spans fails (either some spans are not verbatim, or the union does not entail the target sentence under the entailment check in \cref{fig:prompt-longcite-support}), the prompt is re-issued with the relevant feedback fragment from \cref{fig:prompt-extract-feedback}. We allow up to five retry attempts before falling back to a hard failure.

\begin{figure*}[t]
\footnotesize
\begin{promptbox}{Self-Critique Retry Feedback}
The retry feedback is assembled from three named fragments.
\begin{itemize}
  \item \textbf{Wrapper} (\texttt{EXTRACTION\_FEEDBACK\_SECTION}) wraps the feedback body:
  \begin{quote}\ttfamily
  <feedback>\\
  A previous extraction attempt failed for the following reasons:\\
  \{feedback\_body\}\\
  </feedback>
  \end{quote}
  \item \textbf{Verbatim-match failure} (\texttt{EXTRACTION\_FEEDBACK\_UNFOUND\_SPANS}): Some spans did not appear verbatim in the document they were extracted from. You MUST copy spans exactly as they appear in the source document, character-for-character. \texttt{\{missing\_spans\}}
  \item \textbf{Union-entailment failure} (\texttt{EXTRACTION\_FEEDBACK\_CLAIM\_LEVEL\_FAILED}): The claim-level validator (LongCite \texttt{is\_support}) decision did NOT align with your per-document verdicts:
  \begin{itemize}
    \item Your per-doc \texttt{support\_type}s (most-permissive aggregation across documents): \texttt{[[\{extracted\_strength\}]]} supported.
    \item Validator's decision on the union of spans: \texttt{\{longcite\_decision\}}.
  \end{itemize}
  These must agree on the strength of support. Validator's description of what's missing or in conflict: ``\texttt{\{reasoning\}}''. Either (a)~extract more or better verbatim spans so the union actually reaches the strength your per-doc verdicts claim, (b)~revise per-doc \texttt{support\_type} assignments (e.g., downgrade \texttt{[[Fully]]} to \texttt{[[Partially]]}, or upgrade a \texttt{[[Partially]]} to \texttt{[[Fully]]}) so they reflect what the spans actually support, or (c)~revise \texttt{overall\_decision} to \texttt{[[Not supported]]} if no such spans exist in the documents.
\end{itemize}
\end{promptbox}
\caption{Self-critique retry feedback templates injected into \texttt{\{feedback\_section\}} of \cref{fig:prompt-extract}.}
\label{fig:prompt-extract-feedback}
\end{figure*}

\subsection{Steps 1, 3 and 5: Entailment Test}
\label{appendix:prompt-entailment}

\begin{figure*}[t]
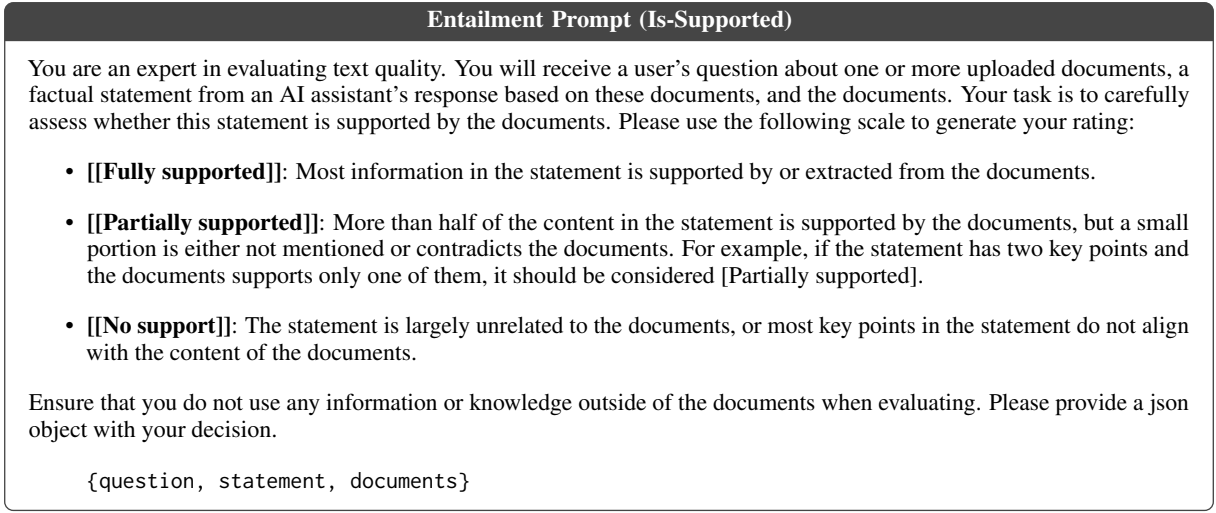

\footnotesize
\begin{promptbox}{Entailment Prompt (Is-Supported)}
You are an expert in evaluating text quality. You will receive a user's question about one or more uploaded documents, a factual statement from an AI assistant's response based on these documents, and the documents. Your task is to carefully assess whether this statement is supported by the documents. Please use the following scale to generate your rating:
\begin{itemize}
  \item \textbf{[[Fully supported]]}: Most information in the statement is supported by or extracted from the documents.
  \item \textbf{[[Partially supported]]}: More than half of the content in the statement is supported by the documents, but a small portion is either not mentioned or contradicts the documents. For example, if the statement has two key points and the documents supports only one of them, it should be considered [Partially supported].
  \item \textbf{[[No support]]}: The statement is largely unrelated to the documents, or most key points in the statement do not align with the content of the documents.
\end{itemize}
Ensure that you do not use any information or knowledge outside of the documents when evaluating. Please provide a json object with your decision.
\begin{quote}\ttfamily
\{question, statement, documents\}
\end{quote}
\end{promptbox}
\caption{Is supported prompt adapted from the LongCite \citep{zhang-etal-2025-longcite}.}
\label{fig:prompt-longcite-support}
\end{figure*}

\Cref{fig:prompt-longcite-support} shows the entailment prompt, adapted from \citet{zhang-etal-2025-longcite} to accept multiple documents and full document texts (rather than chunked passages). This prompt is used for the union-entailment check that validates Step 1, for the entailment tests of Steps 3 and 5, for snippet evaluation (\cref{sec:method}), and for the citation-recall check of \cref{sec:tracing_analysis}.

\begin{figure*}[t]
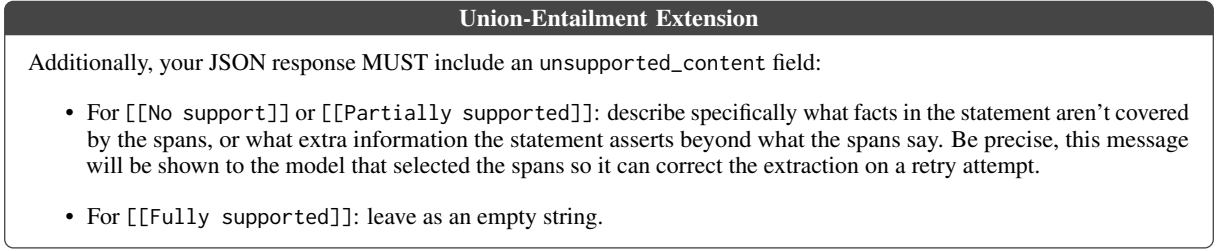

\footnotesize
\begin{promptbox}{Union-Entailment Extension}
Additionally, your JSON response MUST include an \texttt{unsupported\_content} field:
\begin{itemize}
  \item For \texttt{[[No support]]} or \texttt{[[Partially supported]]}: describe specifically what facts in the statement aren't covered by the spans, or what extra information the statement asserts beyond what the spans say. Be precise, this message will be shown to the model that selected the spans so it can correct the extraction on a retry attempt.
  \item For \texttt{[[Fully supported]]}: leave as an empty string.
\end{itemize}
\end{promptbox}
\caption{Extension appended to the prompt in \cref{fig:prompt-longcite-support} for the Step 2 feedback loop.}
\label{fig:prompt-claim-level-extension}
\end{figure*}

For the union check of Step 2, the entailment prompt is composed with the extension in \cref{fig:prompt-claim-level-extension}. The extension requires the model to additionally emit a description what is missing; this description seeds the \texttt{\{reasoning\}} slot of the self-critique feedback (\cref{fig:prompt-extract-feedback}) on the next span-extraction attempt.

\subsection{Steps 2 and 4: Citation Classification}
\label{appendix:prompt-classify}

\Cref{fig:prompt-classify} shows the citation classification prompt, adapted from the DEER citation classification scheme \citep{han-etal-2026-deer}. This prompt is used in Steps 2 and 4 of \cref{sec:method} and in the system-level analysis (\cref{sec:tracing_analysis}) to semantically identify which citations are aligned with a given sentence or span.

\begin{figure*}[t]
\footnotesize
\begin{promptbox}{Citation Classification Prompt}
You are an expert fact-checker. Your task is to \textbf{classify} the provided target sentence from a research report into one of the following categories.

Reports may use any citation style: numbered markers like [1] or [1][3], author-year references like (Kim, 2024), markdown links like [title](url), or other formats. Identify whatever citation style the report uses.

\textbf{Class Definitions:}
\begin{itemize}
  \item \textbf{A: Cited.} Sentence that includes explicit citations (e.g., [1] or (Kim, 2024)) directly within it.
  \item \textbf{B: Uncited -- Same Section/Paragraph.} Sentence whose supporting citations appear within the same section or paragraph. Crucially, the Evidence Position MUST be in the same section or paragraph. If the evidence is in a previous section, it is Class C.
  \item \textbf{C: Uncited -- Previous Section/Paragraph.} Sentence whose supporting citations appear in previous sections or paragraphs. The Evidence Position MUST be in a previous section.
  \item \textbf{D: Uncited -- Recap/Structural.} Simple mentions of introduction, conclusion, abstract, or structural summaries.
  \item \textbf{E: Uncited -- Citation not Required.} Widely accepted facts, common knowledge, author's own findings, subjective claims, methodological contributions, etc. WARNING: Conclusions, results, or findings of the report are NOT Class E. They are specific claims that require evidence (Class A/B/C) or are uncited (Class F).
  \item \textbf{F: No-Citation -- Unknown Source.} Specific factual assertions that require verification but lack any identifiable supporting source in the text. If a claim is specific (e.g., ``X has a value of Y'', ``X is Z\% of GDP'') and has no citation attached to it or the sentence immediately preceding/following it that covers this fact, it is Class F. Do not assume a nearby citation covers it unless it clearly does.
\end{itemize}

\textbf{Instructions:}
\begin{enumerate}
  \item \textbf{Read the Report Context} to understand the global context.
  \item \textbf{Analyze the Target Sentence} and its relationship with the context and citations:
  \begin{itemize}
    \item \textbf{Step 1: Specificity Check.} Contains numbers, statistics, named datasets, specific results? $\rightarrow$ Likely Class A, B, C, or F.
    \item \textbf{Step 2: Citation Check.}
    \begin{itemize}
      \item Citation in same sentence? $\rightarrow$ \textbf{Class A}.
      \item Citation in same paragraph/section? $\rightarrow$ \textbf{Class B}.
      \item Citation in previous section? $\rightarrow$ \textbf{Class C}.
    \end{itemize}
    \item \textbf{Step 3: Uncited Check.}
    \begin{itemize}
      \item Specific fact but no citation? $\rightarrow$ \textbf{Class F}.
      \item General knowledge/methodology/author opinion? $\rightarrow$ \textbf{Class E}.
      \item Structural summary? $\rightarrow$ \textbf{Class D}.
    \end{itemize}
  \end{itemize}
  \item \textbf{Determine \texttt{citation\_identifiers}:}
  \begin{itemize}
    \item For \textbf{Class B} or \textbf{C}, identify the citation identifiers that support the sentence.
    \item For \textbf{Class A}, list the citation identifiers that appear in/around the sentence.
    \item For \textbf{Class D} and \textbf{Class E}, \texttt{citation\_identifiers} must be empty.
    \item Return identifiers exactly as they appear in the report's reference list. For numbered citations like [1], return ``1''. For author-year like (Kim, 2024), return ``Kim, 2024''. For other formats, return the identifier that maps to the source in the reference list.
  \end{itemize}
\end{enumerate}
\end{promptbox}
\caption{Citation classification system prompt, adapted from the DEER citation classification scheme of \citet{han-etal-2026-deer}. The original tabular formatting (a markdown table) is rewritten here as a list for column compatibility, but the content is unchanged. The report and target sentence are provided in a separate user message.}
\label{fig:prompt-classify}
\end{figure*}

\subsection{Citation Recall: Need-Citation Test}
\label{appendix:prompt-need-citation}

For the citation-recall test in \cref{sec:tracing_analysis} (Step 1), we also check whether a sentence requires a citation in the first place. This prompt (\cref{fig:prompt-need-citation}) is adapted from \citet{zhang-etal-2025-longcite}.

\begin{figure*}[t]
\footnotesize
\begin{promptbox}{Need-Citation Prompt}
You are an expert in evaluating text quality. You will receive a user's question regarding their uploaded document (due to the length of the document, it is not shown to you) and an AI assistant's response based on the document. The sentence being evaluated is marked with \texttt{<target>...</target>} tags within the response.

Your task is to determine whether this sentence requires a citation. A sentence DOES NOT require citation if it is:
\begin{itemize}
  \item An introductory sentence that previews information that is comprehensively discussed later in the response (not just briefly mentioned)
  \item A transition sentence connecting ideas
  \item A summary or conclusion based on previous information in the response
  \item Reasoning or inference derived from previous information
\end{itemize}
Ensure that you do not use any other external information during your evaluation. Please provide a json object with your decision (\texttt{[[Yes]]} or \texttt{[[No]]}).
\begin{quote}\ttfamily
\{question, response with <target> tags\}
\end{quote}
\end{promptbox}
\caption{Need-citation prompt, adapted from LongCite \citep{zhang-etal-2025-longcite}.}
\label{fig:prompt-need-citation}
\end{figure*}

\section{Modifications to AI-Q}
\label{appendix:dr-mods}

The application study (\cref{sec:application}) applies two interventions to AI-Q: (1) a prompt change forbidding the use of uncited information by the orchestrator, and (2) replacing AI-synthesized researcher outputs with the raw snippets of the URLs they cite. We describe the implementation below rather than reproducing the full third-party prompts.

\paragraph{AI-Q prompt change.}
A single sentence is appended to the orchestrator's prompt: ``However, do NOT use information which is not backed by citations.''
The sentence is added inside the orchestrator's ``Critical Rules'' section, immediately after the existing instruction to fall back to best-effort analysis when search tools return insufficient data.

\paragraph{AI-Q snippet replacement.}
The researcher subagents normally hand the orchestrator a synthesized note with the raw search-engine snippets of the cited URLs appended at the end. Our intervention drops the synthesized note body and keeps only the appended ``Raw Source Snippets'' section, which is assembled from the citation registry.

\begin{figure*}[t]
\includegraphics[width=\textwidth]{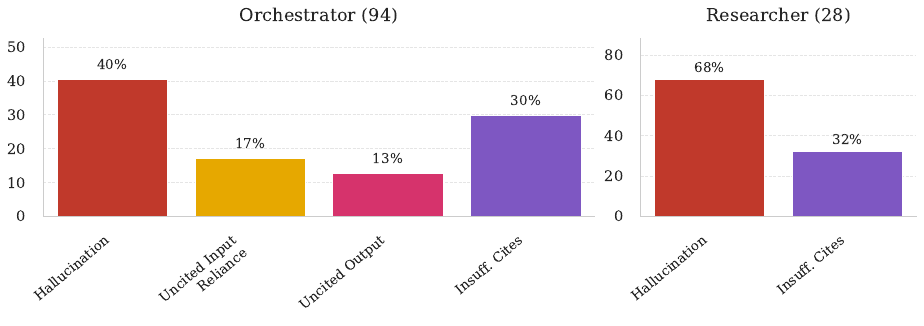}
\centering
\caption{System-level error distribution for Nvidia AI-Q after applying the ``do NOT use information which is not backed by citations.'' prompt change. Compare against \cref{fig:system_level_verdicts}: the orchestrator's \emph{uncited output} bucket is reduced from 48\% to 13\%.}
\label{fig:avoid_uncited_system_level_verdicts}
\end{figure*}

\paragraph{Post-intervention system-level breakdown.}
\Cref{fig:avoid_uncited_system_level_verdicts} shows the full per-agent, per-category system-level error distribution for AI-Q after applying the citation guidance prompt change.  \Cref{tab:avoid_uncited_error_origin} reports the corresponding error-origin distribution. The headline numbers extracted from this figure are discussed in \cref{sec:application}.

\section{Example Errors per Category}
\label{appendix:example_errors}

\Cref{tab:examples} shows representative target sentences flagged by our tracing evaluation (\cref{sec:tracing_analysis}) for each mistake category, with the highlighted portion indicating the problematic span and the explanation describing why the verdict was assigned.

\Cref{tab:error_chains} complements these single-invocation examples with three full propagation chains from the AI-Q runs, showing how an error introduced by one agent invocation interacts with the behavior of the agents that depend on it.

\begin{table*}[t]
\centering
\small
\begin{adjustbox}{max width=\textwidth}
\begin{tabular}{>{\raggedright\arraybackslash}p{0.12\textwidth} >{\raggedright\arraybackslash}p{0.44\textwidth} >{\raggedright\arraybackslash}p{0.44\textwidth}}
\toprule
\textbf{Agent \& Type of error} & \textbf{Target sentence with citations} & \textbf{Explanation} \\
\midrule
Orchestrator -- \newline Hallucination & A Japan-focused cross-sectional study finds \textbf{measurable} consumer willingness-to-pay for food defense and food hygiene, supporting a plausible channel for \textbf{higher value-per-unit food consumption (safer supply chains, trusted brands, convenience-oriented services)}. \newline [https://www.i-jmr.org/2023/1/e43936] & The two highlighted parts of the claim are not supported by the input info units. Interestingly, some of these points are actually discussed in the cited URL, suggesting familiarity of the model with these issues. \\
\midrule
Orchestrator -- \newline Uncited input reliance & \textbf{Attention inspection (when using transformer-like models) can provide a form of structured saliency by showing which time steps or features the model ``attended'' to}; recent portfolio ML frameworks highlight interpretability/efficiency trade-offs using attention structures. [https://www.nature.com/articles/s41598-025-26337-x] & The information highlighted was provided by an uncited researcher synthesis: \newline ``- Attention mechanisms within transformer-style architectures generate saliency maps that highlight which market variables (e.g., volatility, momentum, macro indicators) the network focuses on at each rebalancing step, supporting visual debugging and regime-specific analysis.'' \\
\midrule
Orchestrator -- \newline Uncited output & In 1992 he indicates Berkshire expects to keep most major holdings regardless of how they are priced relative to intrinsic value. & The orchestrator paraphrased a researcher note that cited the Berkshire 1992 shareholder letter but dropped the citation entirely. The researcher's note read: \newline ``We will keep most of our major holdings, regardless of how they are priced relative to intrinsic business value [https://www.berkshirehathaway.com/letters/1992.html].'' \\
\midrule
Orchestrator --\newline Insufficient citations & It also documents tranching and average duration patterns: liquidity tranche around \textbf{10 months} and investment tranche around \textbf{27 months} (with non-tranched portfolios around \textbf{26 months}).\newline[.../Inaugural-RAMP-Survey-on-the-Reserve-Management-Practices-of-Central-Banks-Results-and-Observations.pdf] & The information highlighted was provided by a researcher synthesis with a different World Bank citation: \newline ``Tranching is widely used; the liquidity tranche has an average duration of 10 months, the investment tranche 27 months, while portfolios without tranches average 26 months duration, indicating higher risk tolerance ([.../IDU-f1aeb421-97fb-4a82-a223-0ecbbe0bdbb8.pdf]).''\\
\midrule
Researcher -- \newline Insufficient citations & PwC US announced a three-year, \$1 billion investment to expand and scale its AI capabilities, embedding GPT-class models such as OpenAI's GPT-4/ChatGPT into internal workflows and client solutions; this represents roughly \textbf{0.5\,\% of annual fee income}. \newline [.../pwc-us-makes-billion-investment-in-ai-capabilities.html] & The cited PwC press release supports the \$1 billion investment claim. The highlighted percentage figure is absent from that source and was provided by other input documents (a LinkedIn post and a PwC article) whose citations the researcher did not include. \\
\bottomrule
\end{tabular}
\end{adjustbox}
\caption{Examples for each mistake category produced by our sentence-level evaluation algorithm. }
\label{tab:examples}
\end{table*}

\begin{table*}[t]
\centering
\small
\begin{adjustbox}{max width=\textwidth}
\begin{tabular}{>{\raggedright\arraybackslash}p{0.13\textwidth} >{\raggedright\arraybackslash}p{0.47\textwidth} >{\raggedright\arraybackslash}p{0.40\textwidth}}
\toprule
\textbf{Agent (verdict)} & \textbf{Output text with citations} & \textbf{Explanation} \\
\midrule
\multicolumn{3}{l}{\emph{Chain 1: faithful propagation of a search-tool fabrication (searcher → researcher → orchestrator).}} \\
\midrule
Searcher -- \newline Hallucination & The United States Space Force has formalized its cislunar effort by creating a Cislunar Coordination Office [...] officials have said the office will serve as the central hub for coordinating cislunar programs across the service and \textbf{will ensure that space-situational-awareness data [...] is shared with NASA, other Department of Defense components, allied partners, and civilian space stakeholders}, and this commitment was underscored by [...] the Oracle-M hot-fire test [...] which \textbf{confirmed that the resulting data will be integrated into joint databases and disseminated to national-security and civil-space users}. & The searcher's answer summarizes the pages retrieved for the query, but the highlighted data-sharing and database-integration commitments appear in none of them. \\
Researcher -- \newline Inherited & The office will integrate SSA data into \textbf{joint databases} and disseminate the information to both national-security and civil-space users to enhance safety, security, and mission planning. \newline [.../space-force-sets-up-office-to-coordinate-cislunar-programs/] [.../oracle-m-hot-fire-test-a-major-milestone-in-cislunar-space-situational-awarenes] & The researcher's note restates the fabricated commitment and attaches the URLs that the searcher provided for this claim. \\
Orchestrator -- \newline Inherited & Public reporting indicates the U.S. Space Force has stood up a cislunar coordination function and is advancing systems (e.g., Oracle-M) aimed at monitoring and tracking cislunar objects, \textbf{with plans to integrate tracking data into shared SSA databases}. \newline [.../space-force-sets-up-office-to-coordinate-cislunar-programs/] [.../oracle-m-hot-fire-test-a-major-milestone-in-cislunar-space-situational-awarenes] [.../space-force-sets-up-cislunar-coordination-office-to-focus-beyond-earth-orbit/] & The final-report sentence faithfully summarizes the researcher's notes and propagates their citations, so the orchestrator's local test passes as well. The fabricated claim reaches the reader carrying citations to real articles that do not state the information. \\
\midrule
\multicolumn{3}{l}{\emph{Chain 2: abstraction masks a researcher fabrication (researcher → orchestrator).}} \\
\midrule
Researcher -- \newline Hallucination & Then enable scheduled autoscaling for the cluster (\textbf{
\texttt{gcloud container clusters update CLUSTER\_NAME -{}-enable-scheduled-autoscaling}}). Create or modify a schedule using \texttt{gcloud container node-pools update POOL\_NAME -{}-add-chedule=start\_time=HH:MM,\allowbreak end\_time=HH:MM,\allowbreak min-nodes=MIN,\allowbreak max-nodes=MAX} (or via console). The schedule can be daily or weekly. \newline [.../kubernetes-engine/docs/concepts/cluster-autoscaler] & The researcher invents an entire GKE ``scheduled autoscaling'' workflow, including command-line flags that do not appear in the cited official documentation, and attaches the real cluster-autoscaler documentation as its citation. \\
Orchestrator -- \newline Inherited & GKE: scheduled bounds on node pools via managed capabilities; ensure CA is enabled and schedules reflect business windows. \newline [.../kubernetes-engine/docs/concepts/cluster-autoscaler] [.../kubernetes-engine/docs/how-to/cluster-autoscaler] & The orchestrator abstracts the fabricated procedure into a hedged, higher-level claim that is fully supported by the researcher's notes. The abstraction propagates the fabrication, masks it behind vaguer language, and keeps its citations. \\
\midrule
\multicolumn{3}{l}{\emph{Chain 3: unsupported statistics travel verbatim with their citation (researcher → orchestrator).}} \\
\midrule
Researcher -- \newline Insufficient citations & According to the OECD 2023 Annual Survey, infrastructure represents roughly \textbf{6--7\,\%} of assets for large public-pension-related investors, with public pension reserve funds allocating about \textbf{9\,\%} of their portfolios to infrastructure. \newline [.../long-term-investing-of-large-pension-funds-and-public-pension-reserve-funds-2023\_042fb731/c690ccc3-en.pdf] & The highlighted figures were available to the researcher from other input documents, but the note cites only the OECD report, which does not support them. \\
Orchestrator -- \newline Inherited & The OECD 2023 survey reports infrastructure at roughly \textbf{6--7\%} of assets for large public-pension-related investors and about \textbf{9\%} for public pension reserve funds. \newline [.../long-term-investing-of-large-pension-funds-and-public-pension-reserve-funds-2023\_042fb731/c690ccc3-en.pdf] & The figures and their citation travel together across the compression boundary into the report. Because the orchestrator's sentence is faithful to the researcher note it cites, only recursing into the researcher reveals that the attached citation never supported the figures. \\
\bottomrule
\end{tabular}
\end{adjustbox}
\caption{Representative error-propagation chains produced by our system-level tracing evaluation on AI-Q, illustrating how error types interact across agents.}
\label{tab:error_chains}
\end{table*}

\section{Human Annotation Guidelines}
\label{appendix:annotation_guidelines}

\Cref{fig:annotation-guidelines} includes verbatim the guidelines given to the human annotators whose agreement with our algorithm is reported in \cref{sec:human_analysis}. Each annotation task showed the annotator the target sentence (in a Report tab) and the input info units available to the writing agent (in an ``Information providers'' tab), and asked them to pick one of six labels. Over an initial online orientation meeting, each annotator completed two calibration tasks before beginning the main annotation, and annotators were informed that their annotations would be used and published as part of an academic research project. 

Labels 2--6 correspond directly to the algorithm's terminal verdicts in \cref{sec:synthesized}: \emph{hallucination}, \emph{uncited input reliance}, \emph{uncited output}, \emph{insufficient citations}, and \emph{correct}. Label 1 (N/A) corresponds to sentences that the algorithm filters out via the need-citation prompt of \cref{appendix:prompt-need-citation}.

\begin{figure*}[t]
\footnotesize
\begin{promptbox}{Annotation guidelines}
Your task is to determine whether the target sentence (extracted from a deep research
report -- see the Report tab) is faithful to and correctly cited against the
information providers (see the Information providers tab). \\

\textbf{Information providers -- three types.}\\ Three kinds of providers may appear in the
Information providers tab. The deep research report's writer may extract or paraphrase
information from any of them. \\
  1. Planner -- a JSON plan from the agent's planner subagent. \\
  2. Researcher -- a short ~1-page research note from a researcher subagent. \\
  3. Verified-sources registry -- a numbered list of all URLs the agent's subagents
     verified. \\
\\
\textbf{Planner provider (exactly one) will usually not have citations.} Researcher providers
(more than one) usually carry their own citations. Each researcher provider has its
own References / Sources section with its own [N] numbering -- don't conflate citation
indices between different providers. Verified-sources registry (exactly one) is also
important because the target sentence might use information based on their titles.
Check now the providers tab and make sure you can find the different providers and
their citations. \\
\\
\textbf{Citations may live elsewhere.} When deciding which citations belong to a sentence, look
at both inline [N] markers and the surrounding context -- an earlier sentence/
paragraph, or a "References" / "Sources" block at the end. \\
\\
\textbf{Clarification -- no need to open external URLs.} Throughout the annotation process
there is no need to open external URLs. All questions are about the correctness of the
report with respect to the text and citations in the information providers tab. \\
\\
\textbf{Searching tip.} When looking for sentences that support the target, use Ctrl+F and try
paraphrases of the target's key terms -- the writer may have paraphrased the
underlying source, so an exact-word match isn't guaranteed. Note: in Google Docs,
Ctrl+F may jump across tabs -- always check which tab you're currently viewing. \\
\\

\textbf{Label definitions} \\
Your task is to select the most appropriate label for each target sentence based on
its faithfulness and citation accuracy. \\
  1. N/A (doesn't need citations) -- the target does not need citations. \\
  2. Hallucination or partial hallucination -- providers do not fully support the
     target. \\
  3. Uncited input reliance -- providers do support the target, but one or more of the
     provider sentences necessary to support the target do not have citations. \\
  4. Uncited output -- provider sentences that are needed to support the target have
     citations, but the target sentence has no citations at all. \\
  5. Insufficient citations -- provider sentences that are needed to support the
     target have citations, but the target's citations don't cover those provider
     citations. \\
  6. No issues (correct) -- the target is supported and correctly cited. \\
\\
\textbf{Note.} It's possible that there is more than one provider sentence that supports the
target. If you chose any label other than "No issues (correct)", make sure you didn't
miss another possible supporting sentence in the providers.
\end{promptbox}
\caption{Verbatim copy of the human annotation guidelines shown to annotators (see \cref{sec:human_analysis}). The per-task worked example is omitted for space. The task was presented in a google doc with three tabs: the guidelines presented above, the output that the agent created, and the inputs that the agent received.}
\label{fig:annotation-guidelines}
\end{figure*}

\begin{figure*}[t]
\includegraphics[width=\textwidth]{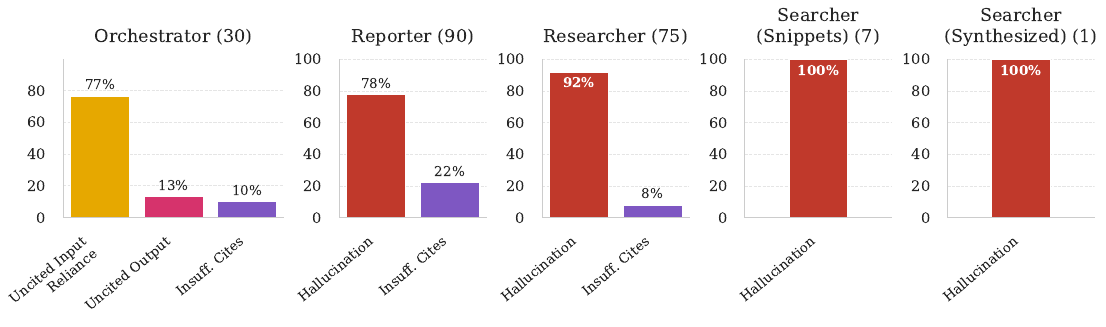}
\centering
\caption{Agent-level error distribution for MS-Agent. The counterpart for Nvidia AI-Q is \cref{fig:agent_level_analysis}.}
\label{fig:ms_agent__agent_level_analysis}
\end{figure*}

\begin{figure*}[t]
\includegraphics[width=\textwidth]{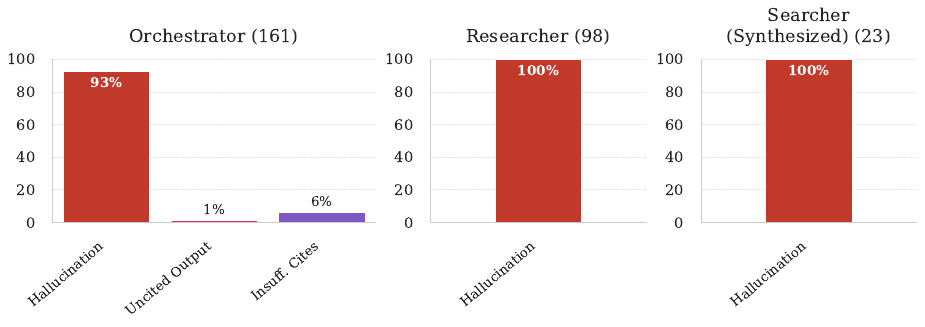}
\centering
\caption{Agent-level error distribution for TrajectoryKit. Unlike the other two systems, hallucinations dominate at every agent rather than giving way to citation errors deeper in the information flow. The counterpart for Nvidia AI-Q is \cref{fig:agent_level_analysis}.}
\label{fig:trajectorykit__agent_level_analysis}
\end{figure*}

\begin{figure*}[t]
\includegraphics[width=\textwidth]{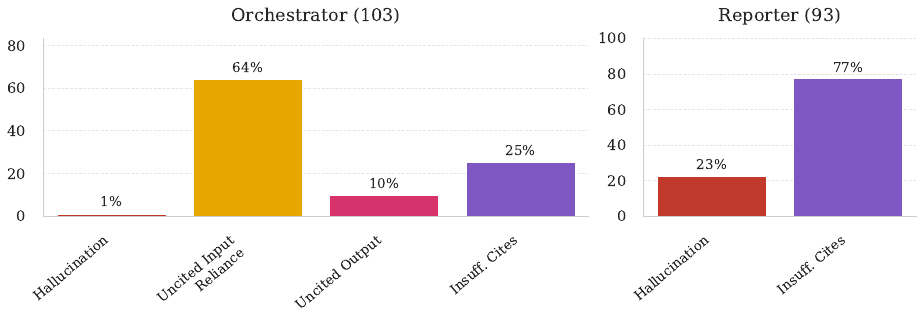}
\centering
\caption{System-level error distribution for MS-Agent. The counterpart for Nvidia AI-Q is \cref{fig:system_level_verdicts}.}
\label{fig:ms_agent__system_level_verdicts}
\end{figure*}

\begin{figure}[t]
\includegraphics[width=\columnwidth]{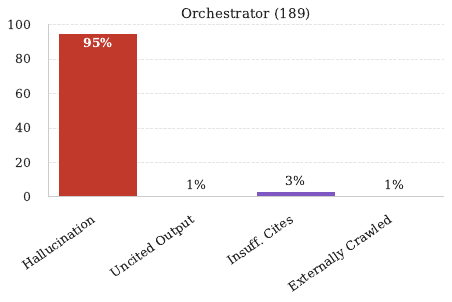}
\centering
\caption{System-level error distribution for TrajectoryKit. Every confirmed final-report error is attributed to the orchestrator, and 95\% of them are hallucinations. The counterpart for Nvidia AI-Q is \cref{fig:system_level_verdicts}.}
\label{fig:trajectorykit__system_level_verdicts}
\end{figure}

\section{Validating the Supporting Spans}
\label{appendix:span_validation}

\Cref{sec:human_analysis} validates the \emph{classification} produced by the local test. This appendix validates its other output, the \emph{supporting spans}: the input spans the algorithm identifies as supporting a target sentence.

\paragraph{Why these two outputs suffice.}
The system-level tracing of \cref{sec:tracing_analysis} adds no model-based judgment beyond the local test. Each recursive step reuses the supporting spans that the local algorithm of \cref{sec:method} already produced: the spans determine which agent invocation to descend into (\cref{sec:tracing_algo}), and the descent itself involves no additional model calls. The error-prone components of the pipeline are therefore the local test's classification and its supporting spans, so validating both directly validates the system-level tracing built on top of them.

\paragraph{Annotation.}
In the same human evaluation described in \cref{sec:human_analysis}, annotators also recorded which input spans they used to support their classification. For each such sentence we compare the input spans marked by the annotators against those returned by the localization process, counting them as agreeing when they select the same supporting input sentences.

\paragraph{Results.}
For $75\%$ of the relevant target sentences, the localization process agrees with the annotators on the input sentences that support the target sentence. As expected, this closely tracks the $76\%$ exact-match rate on the classification task reported in \cref{sec:human_analysis}, indicating that the two outputs of the local test are reliable to a similar degree.

\section{More Experimental Details}

DeepResearch Bench \citep{du-etal-2025-drb} spans 22 domains and includes questions in both English and Chinese. In our experiments, we used only English examples and leave multilingual evaluation to future work. The diagnostic analyses (\cref{sec:agent_analysis,sec:tracing_analysis}) use 20 examples (numbered 51--70), and the intervention study (\cref{sec:application}) extends this to all 50 English examples (numbered 51--100).

The LLM model used as the judge is \texttt{gpt-5-mini-2025-08-07} \citep{singh2026openaigpt5card}.

\begin{table}[t]
\centering
\small
\begin{tabular}{lll}
\toprule
\textbf{System} & \textbf{Agent} & \textbf{LLM} \\
\midrule
\multirow{5}{*}{Nvidia AI-Q}
 & Orchestrator    & GPT-5.2 \\
 & Planner         & GPT-5.2 \\
 & Researcher      & Nemotron Nano 30B \\
 & Tavily Snippets & (Tavily API) \\
 & Tavily Answer   & (Tavily API) \\
\midrule
\multirow{4}{*}{MS-Agent}
 & Orchestrator & GPT-5 \\
 & Reporter     & GPT-5 mini \\
 & Researcher   & GPT-5 mini \\
 & Searcher     & GPT-5 nano \\
\midrule
\multirow{3}{*}{TrajectoryKit}
 & Orchestrator       & gpt-oss-20b \\
 & Worker             & gpt-oss-20b \\
 & Summarize Webpage  & gpt-oss-20b \\
\bottomrule
\end{tabular}
\caption{LLM used by each agent in the three open-source DR systems we evaluate. Nemotron Nano 30B refers to \citet{nvidia_nemotron_nano_v3_2025}. Tavily Snippets and Tavily Answer are calls to a search API rather than direct prompts to an LLM.}
\label{tab:agent_llms}
\end{table}

\section{Licenses}
\label{appendix:licenses}

All artifacts used in this work are released under permissive open-source licenses, and our use is consistent with their intended use.

\begin{itemize}
\item \textbf{DeepResearch Bench} \citep{du-etal-2025-drb}: both the codebase and the accompanying dataset are released under the Apache License 2.0.
\item \textbf{Nvidia AI-Q} \citep{nvidia2026aiq}: released under the Apache License 2.0.
\item \textbf{MS-Agent} \citep{modelscope2026msagent}: released under the Apache License 2.0.
\item \textbf{TrajectoryKit} \citep{trajectorykit2026}: released under the MIT License.
\end{itemize}

\section{System-specific Logic}\label{appendix:system_specific_logic}

Our evaluation framework (\cref{sec:framework}) requires extracting the inputs and outputs of each agent invocation from the system's execution trace. Because AI-Q, MS-Agent and TrajectoryKit differ in their trace formats, agent organization, and citation conventions, each system requires a dedicated \emph{adapter} that maps the raw trace into the shared data structures consumed by our algorithm. Despite these differences, the evaluation algorithm itself (\cref{sec:algorithm}) runs identically on all three systems: each adapter only needs to expose, per agent, the inputs and outputs. Below we describe the system-specific extraction logic; \Cref{tab:per-system-io} provides a summary.

\paragraph{Agent naming.}
The agent names used in each system's codebase differ from the abstract roles used in this paper (\cref{fig:dr_systems_information_flow}). In AI-Q, what we call the ``researcher'' is internally named \texttt{searcher-agent}, and what we call the ``searcher'' corresponds to the Exa Search + Jina + Content Optimizer module. In MS-Agent, the orchestrator role is split between a \texttt{reporter\_tool} subagent (which writes the initial report draft) and the top-level orchestrator (which issues post-edits); our adapter attributes errors to the specific phase that introduced them.  In TrajectoryKit, what we call a ``researcher'' is a \texttt{conduct\_research} subagent (internally a \emph{worker}), and what we call its ``searcher'' is the orchestrator's \texttt{summarize\_webpage} tool, which produces the synthesized per-page digests.

\paragraph{The searcher differs significantly between systems.}
In AI-Q the searcher is the \texttt{<Answer>} block produced by the Tavily search API itself (\texttt{include\_answer="advanced"}) and summarizes all documents returned for one query.  In MS-Agent it is a single user message, with no system prompt, sent to a separate, smaller model (\texttt{gpt-5-nano}) by the content optimizer inside the web-search tool, once per document, over at most 50K characters, and skipped entirely for pages shorter than 2K characters. In TrajectoryKit it is a system and user message pair sent to the same \texttt{gpt-oss-20b} that the agents themselves run on, once per document, over at most 200K characters, from inside the orchestrator's \texttt{summarize\_webpage} handler. We nevertheless evaluate these nodes exactly as we evaluate agents, because each performs the same compression of retrieved documents whose citation behavior our framework measures.

\subsection{AI-Q}

\paragraph{Trace parsing and event attribution.} Our implementation adds code to to produce a flat sequence of numbered JSON event files (\texttt{0000\_*.json}, \texttt{0001\_*.json}, \ldots), each tagged with a \texttt{kind} (\texttt{tool\_call}, \texttt{tool\_result}, or \texttt{llm}) and an \texttt{agent} identifier. This facilitates the analysis of the outputs.

\paragraph{Orchestrator inputs.}
The orchestrator's inputs are all returned subagent outputs plus the source registry. Subagent outputs are recovered via two channels: the \texttt{task} tool result (the subagent's final message) and orchestrator \texttt{read\_file} calls on files the subagent wrote. Our adapter tracks both and attributes each file read to the subagent that last wrote the requested path.

\paragraph{Researcher inputs.}
Each researcher's inputs are the web-search tool responses it received, which contain XML-style \texttt{<Document href="...">} blocks (verbatim snippets from source pages) and \texttt{<Answer>} blocks (synthesized by the Tavily search API). Our adapter splits these into separate evaluation targets: \texttt{<Document>} snippets are evaluated as extracted content (\cref{sec:snippets}), while \texttt{<Answer>} blocks are evaluated as synthesized content (\cref{sec:synthesized}).

\paragraph{Searcher outputs.}
The web-search tool is Tavily, configured to return the top three results for a query together with an answer field (\texttt{include\_answer="advanced"}). The \texttt{<Document>} blocks are the per-result page content the service returns, and the \texttt{<Answer>} block is that answer field, produced once per query over the results returned for it. It is the only synthesized-searcher output among the three systems that covers several documents at once rather than a single page.

\paragraph{Planner preprocessing.}
AI-Q's planner emits a JSON plan rather than prose. Because our span-extraction and entailment prompts expect natural language, the adapter converts the JSON to Markdown: keys become section headers and values become body text. This conversion handles three observed JSON shapes (whole-text JSON, fenced code blocks, and concatenated JSON regions).

\paragraph{Source registry.}
The orchestrator can call \texttt{get\_verified\_sources}, returning a numbered list of all URLs verified during the run (formatted as \texttt{[N] Title: URL}). Each entry is treated as an extracted snippet, enabling the algorithm to detect when the orchestrator cites a URL based solely on its registry title rather than on full document content.

\subsection{MS-Agent}

\paragraph{Trace parsing.}
For MS-Agent we produce a numbered event stream similar to AI-Q.

\paragraph{Reporter and orchestrator split.}
MS-Agent's final report is produced in two phases: a reporter subagent writes the initial draft via \texttt{write\_file}, and the orchestrator issues incremental \texttt{replace\_file\_contents} edits. 

\paragraph{Evidence note architecture.}
MS-Agent's searchers produce structured evidence notes, each with a unique hex ID, content sections (title, content, contradicting evidence, summary), and a metadata list of source URLs with quality tiers. Downstream agents cite these notes using \texttt{[Note: <hex>]} markers rather than standard URL citations.

\paragraph{Researcher synthesis.}
Each researcher's final response to the orchestrator contains a JSON synthesis with fields for status, task summary, findings, and a report. This synthesis cites the notes the searcher authored using \texttt{[Note: <hex>]} markers.

\paragraph{Searcher outputs.}
MS-Agent's search pipeline uses Exa search followed by a content optimizer that produces a summary and key excerpts per URL. The optimizer runs inside the web-search tool as a single prompt to a separate, smaller model, given the page content alone and truncated to 50K characters; pages shorter than 2K characters bypass it entirely and are carried through verbatim in place of a summary. Key excerpts are evaluated as extracted snippets (\cref{sec:snippets}), while summaries are evaluated as synthesized content (\cref{sec:synthesized}).

\subsection{TrajectoryKit}

\paragraph{Trace parsing and event attribution.} For TrajectoryKit we produce a numbered event stream similar to AI-Q.

\paragraph{Orchestrator inputs.}
The root orchestrator is given a restricted, virtual tool set: it can delegate a research task (\texttt{conduct\_research}), rewrite its draft (\texttt{refine\_draft}), re-read earlier drafts, publish the draft (\texttt{research\_complete}), think, and summarize a single known URL (\texttt{summarize\_webpage}). It never searches or fetches documents itself. Its inputs are therefore the final responses of its researcher subagents, delivered as \texttt{conduct\_research} tool results, together with the page digests returned by its own \texttt{summarize\_webpage} calls. Unlike MS-Agent, there is no shared file system and no source registry, so this tool-result channel is the only route by which subagent content reaches the orchestrator.

\paragraph{Researcher inputs.}
Each \texttt{conduct\_research} subagent receives a fresh context and the full tool set: web search plus the fetch tools \texttt{fetch\_url}, \texttt{read\_page}, \texttt{read\_pdf}, \texttt{extract\_tables} and \texttt{fetch\_cached}. The researcher's inputs are the tool responses it received. A researcher's output does not have \texttt{[N]} markers, so it is attributed structurally to the URLs that researcher surfaced through its own search and fetch results.

\paragraph{Searcher outputs.}
Searcher output takes two forms. The first, \texttt{search\_web}, is Google through the Serper API, returning a numbered listing of title, URL and snippet per result. We do not treat these entries as an evaluation target, because unlike the snippet leaves of AI-Q and MS-Agent we don't have access to the raw document that the system retrieved, so the test that scores extracted content has no reliable reference. The second form is the orchestrator's \texttt{summarize\_webpage} tool, which fetches one URL and runs a single prompt over its content, truncated to 200K characters, on the same \texttt{gpt-oss-20b} the agents themselves run on, optionally steered by a \texttt{focus} argument.

\paragraph{Citation format.}
The orchestrator's draft is a living document that each \texttt{refine\_draft} call replaces in full, and \texttt{research\_complete} publishes the latest version. On publication, TrajectoryKit rewrites the \texttt{[N]} markers in the report body into Markdown links of the form \texttt{[[N]](url)}, resolved against the report's \texttt{\#\# Sources} section (formatted as \texttt{[N] Title: URL}). Our adapter normalizes these links, and the full-width-bracket variant the model occasionally emits, back to plain \texttt{[N]} markers, after which the report format is identical to AI-Q's and the same reference-section parsing applies.

\paragraph{Absence of a planner.}
TrajectoryKit's closest analogue to AI-Q's planner is a single chain-analysis call issued before the orchestrator loop starts, which decomposes the question into ordered subtasks. Because it runs before any retrieval, it consumes no documents and produces no citations, so it is not an evaluation target and TrajectoryKit contributes no planner row to our analyses.

\paragraph{Fallback writer.}
If the orchestrator exhausts its turn budget without publishing, TrajectoryKit writes the report outside the main loop: first by forcing a final answer within the orchestrator's own context, and, failing that, by spawning a dedicated synthesizer subagent over the accumulated research data. We attribute the final report to the orchestrator in either case.

\subsection{Citation Attribution Modes}\label{appendix:citation_modes}

Agents use one of three citation attribution modes depending on their output format:
\begin{itemize}
    \item \emph{Semantic}: citations are identified per-sentence using the DEER classification scheme (\cref{appendix:prompt-classify}).
    \item \emph{Structural}: citations are inherited from the structural context. For example, the source URLs of a referenced note.
    \item \emph{Extracted}: the output is a verbatim excerpt of a single source, so the citation is the identity of the page it was taken from. Used for the searcher snippet leaves of AI-Q and MS-Agent, whose claims are checked by a verbatim test against the full crawled document rather than by an entailment test.
\end{itemize}

The mode follows from the output format rather than from the agent's position in the information flow, so the same role can be attributed differently across systems. For example, AI-Q's researcher notes carry explicit \texttt{[N]} markers that resolve to URLs, so they can be classified per sentence. An MS-Agent evidence note instead lists its sources as note metadata rather than inline, so each of its sentences inherits that note's source list. Similarly, in TrajectoryKit there are no inline markers, so sentences inherit the union of the URLs that researcher surfaced through its own search and fetch calls. Structural attribution is the conservative choice in both cases: crediting a sentence with every source its author had at hand errs toward finding support rather than toward flagging a citation error.

\begin{table*}[t]
\centering
\small
\begin{tabular}{llp{4.6cm}p{3.4cm}l}
\toprule
\textbf{System} & \textbf{Agent} & \textbf{Inputs (for span extraction)} & \textbf{Outputs (evaluation targets)} & \textbf{Cit.\ mode} \\
\midrule
\multirow{5}{*}{AI-Q}
 & Orchestrator & Researcher notes, planner plans, source registry & Final report & Semantic \\
 & Researcher & \texttt{<Document>} snippets + \texttt{<Answer>} blocks & Synthesized note with {[N]} cit. & Semantic \\
 & Planner & Same as Researcher & JSON plan $\to$ Markdown & - \\
 & Searcher (snippets) & Raw crawled documents & \texttt{<Document>} blocks & Extracted \\
 & Searcher (answers) & Raw crawled documents & \texttt{<Answer>} blocks & Structural \\
\midrule
\multirow{5}{*}{MS-Agent}
 & Orchestrator & Reporter snapshots, analyses, conflicts & Final report (post-edit) & Semantic \\
 & Reporter & Syntheses, notes, analyses & Report draft & Semantic \\
 & Researcher (notes) & Summaries + key excerpts per URL & Evidence notes & Structural \\
 & Searcher (synthesis) & Raw crawled documents & Synthesis JSON & Structural \\
 & Searcher (snippets) & Raw crawled documents & Synthesis JSON & Extracted \\
\midrule
\multirow{3}{*}{TrajectoryKit}
 & Orchestrator & Researcher final responses, \texttt{summarize\_webpage} digests & Final report & Semantic \\
 & Researcher & Search snippets + fetched page bodies & Final response to orchestrator & Structural \\
 & Searcher (answers) & One fetched page & \texttt{summarize\_webpage} digest & Structural \\
\bottomrule
\end{tabular}
\caption{Per-agent extraction summary. \emph{Inputs} are the agent invocation inputs provided to the span-extraction step of our algorithm (\cref{sec:algorithm}); \emph{outputs} are the evaluation targets; \emph{citation mode} indicates how citations are identified (semantic = per-sentence DEER classification; structural = inherited from context; extracted = source URL identity).}
\label{tab:per-system-io}
\end{table*}

\subsection{The Citation of Notes in MS-Agent}

Unlike AI-Q, which uses standard \texttt{[N]} citation markers that map directly to URLs, MS-Agent's intermediate outputs cite structured \emph{evidence notes} using markers such as \texttt{[Note: 02e74b]}. Each note is a structured object with content sections and a list of source URLs with quality tiers.
For example, a searcher synthesis might contain:
\begin{quote}
\small\texttt{Berkshire stresses conservative leverage, large cash/float cushions, and a decentralized structure trusting operating managers. [Note: 02e74b]}
\end{quote}
These note markers are not standard URL citations and would be meaningless to end users reading the final report. Notes in turn have structured source metadata (e.g., \texttt{\{"title": ..., "sources": [\{"url": "..."\}]\}}).

Our adapter handles note citations differently depending on their position in the information flow:

\paragraph{Intermediate agents.} For the reporter and searcher synthesis, citing a note is considered acceptable. The downstream orchestrator has access to the notes and their source URLs, and is expected to resolve note citations into proper URL citations before producing the final report.

\paragraph{Final report.} If the final report cites a note marker, we drop this citation because it carries no meaning for the reader. This could result in citation-related errors if the note did support the content.

\paragraph{Conflict objects.} MS-Agent’s reporter may also cite \emph{conflict objects} using \texttt{[conflict\_<hex>]} markers. Each conflict records a discrepancy between sources with a description, resolution, and list of evidence note IDs. The adapter resolves conflict citations through the referenced notes’ source URLs, adding one additional level of indirection to the resolution chain.

\end{document}

%% file: auto_generated_figures/agent_level_performance.tex
\begin{tabular}{clcrr}
\toprule
\textbf{} & \textbf{Component} & \textbf{\#} & \textbf{Correct \%} & \textbf{Mistakes \%} \\
\midrule
\multirow{5}{*}{\rotatebox[origin=c]{90}{AI-Q}} & Orchestrator & 193 & 69.1 & 30.9 \\
 & Planner$^{*}$ & 121 & 78.0 & 22.0 \\
 & Researcher & 200 & 29.2 & 70.8 \\
 & Searcher (Synthesized) & 192 & 38.0 & 62.0 \\
 & Searcher (Snippets) & 189 & 96.2 & 3.8 \\
\midrule
\multirow{5}{*}{\rotatebox[origin=c]{90}{MS-Agent}} & Orchestrator & 183 & 83.2 & 16.8 \\
 & Reporter & 190 & 38.4 & 61.6 \\
 & Researcher & 170 & 53.4 & 46.6 \\
 & Searcher (Synthesized) & 110 & 99.1 & 0.9 \\
 & Searcher (Snippets) & 110 & 93.6 & 6.4 \\
\midrule
\multirow{3}{*}{\rotatebox[origin=c]{90}{TrajKit}} & Orchestrator & 176 & 7.5 & 92.5 \\
 & Researcher & 200 & 50.0 & 50.0 \\
 & Searcher (Synthesized) & 154 & 85.1 & 14.9 \\
\bottomrule
\end{tabular}

%% file: auto_generated_figures/system_level_error_origin_distribution.tex
\begin{tabular}{llr}
\toprule
\textbf{System} & \textbf{Component} & \textbf{\% of errors} \\
\midrule
\multirow{3}{*}{AI-Q} & Orchestrator & 84.7 \\
 & Researcher & 14.8 \\
 & Searcher (Synthesized) & 0.4 \\
\midrule
\multirow{2}{*}{MS-Agent} & Orchestrator & 52.6 \\
 & Reporter & 47.4 \\
\midrule
\multirow{1}{*}{TrajectoryKit} & Orchestrator & 100.0 \\
\bottomrule
\end{tabular}

%% file: auto_generated_figures/aiq_intervention.tex
\begin{tabular}{lccc}
\toprule
\textbf{System} & \textbf{RACE} & \textbf{C. Recall} & \textbf{C. Precision} \\
\midrule
AI-Q & 52.6\tiny{$\pm{0.4}$} & 64.5\tiny{$\pm{3.9}$} & 87.6\tiny{$\pm{3.0}$} \\
\quad{}+ replace w/ snippets & 52.4\tiny{$\pm{0.7}$} & \textbf{69.7\tiny{$\pm{4.1}$}} & \textbf{94.1\tiny{$\pm{2.0}$}} \\
\quad{}+ citation guidance & 52.6\tiny{$\pm{0.6}$} & \textbf{69.6\tiny{$\pm{3.2}$}} & 91.0\tiny{$\pm{2.3}$} \\
\bottomrule
\end{tabular}

%% file: auto_generated_figures/system_level_error_origin_distribution_avoid_uncited.tex
\begin{tabular}{llr}
\toprule
\textbf{System} & \textbf{Component} & \textbf{\% of errors} \\
\midrule
\multirow{2}{*}{AI-Q + citation guidance} & Orchestrator & 77.0 \\
 & Researcher & 23.0 \\
\bottomrule
\end{tabular}